\documentclass[10pt,twocolumn,letterpaper]{article}

\usepackage[pagenumbers]{iccv}      % To produce the REVIEW version
\definecolor{iccvblue}{rgb}{0.21,0.49,0.74}
\usepackage[pagebackref,breaklinks,colorlinks,allcolors=iccvblue]{hyperref}
\usepackage{svg}
\usepackage{tcolorbox}
\usepackage{multirow}
\usepackage{bm}
\usepackage{fontawesome}
\usepackage{hyperref}

\def\paperID{6112} % *** Enter the Paper ID here
\def\confName{ICCV}
\def\confYear{2025}

\title{Bilateral Collaboration with Large Vision-Language Models for\\ Open Vocabulary Human-Object Interaction Detection}

\author{
    Yupeng Hu$^1$ \quad 
    Changxing Ding$^{1}$\footnotemark[2] \quad
    Chang Sun$^1$ \quad 
    Shaoli Huang$^2$  \quad 
    Xiangmin Xu$^{1,3}$\\
    $^1$South China University of Technology\quad 
    $^2$Tencent AI Lab \quad
    $^3$Foshan University \\
{\tt\small \{ee202320111334, eesunchang2024\}@mail.scut.edu.cn, shaol.huang@gmail.com} \\
{\tt\small \{chxding, xmxu\}@scut.edu.cn}}
\begin{document}
\maketitle
\footnotetext[2]{Corresponding author.}
\begin{abstract}

Open vocabulary Human-Object Interaction (HOI) detection is a challenging task that detects all $<$human, verb, object$>$ triplets of interest in an image, even those that are not pre-defined in the training set. Existing approaches typically rely on output features generated by large Vision-Language Models (VLMs) to enhance the generalization ability of interaction representations. However, the visual features produced by VLMs are holistic and coarse-grained, which contradicts the nature of detection tasks. To address this issue, we propose a novel \textbf{B}ilateral \textbf{C}ollaboration framework for open vocabulary \textbf{HOI} detection (BC-HOI). This framework includes an Attention Bias Guidance (ABG) component, which guides the VLM to produce fine-grained instance-level interaction features according to the attention bias provided by the HOI detector. It also includes a Large Language Model (LLM)-based Supervision Guidance (LSG) component, which provides fine-grained token-level supervision for the HOI detector by the LLM component of the VLM. LSG enhances the ability of ABG to generate high-quality attention bias. We conduct extensive experiments on two popular benchmarks: HICO-DET and V-COCO, consistently achieving superior performance in the open vocabulary and closed settings. Code is available at \href{https://github.com/MPI-Lab/BC-HOI}{https://github.com/MPI-Lab/BC-HOI}.

\begin{figure}[htp]
    \centering
    \includegraphics[width=\linewidth]{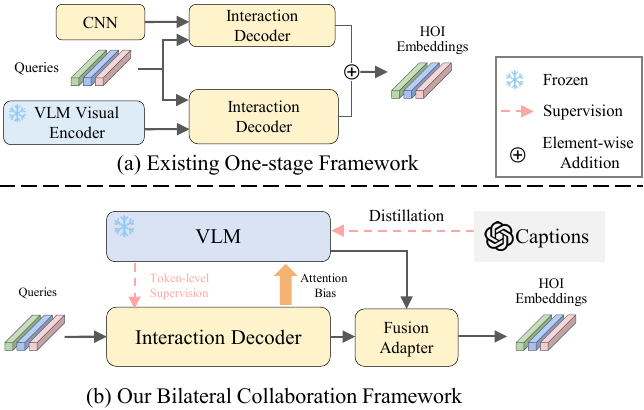}
    \caption{Comparisons between one-stage open vocabulary HOI detection frameworks. (a) Existing works \cite{2023hoiclip,unihoi} usually utilize HOI queries to extract interaction features from the CNN backbone and VLM, respectively. Then, they fuse these features via element-wise addition. (b) Our approach establishes bilateral collaboration between the VLM and HOI detector. The HOI detector provides attention bias to the VLM, enabling the VLM to extract fine-grained interaction features. The VLM provides fine-grained token-level supervision to the HOI detector, enabling the latter to generate high-quality attention bias.}
    \label{fig:intro1}
\end{figure}

\end{abstract}    
\section{Introduction}
\label{sec:intro}

Human-Object Interaction (HOI) detection \cite{unihoi-15} is a human-centric scene understanding task that localizes all the interacting human-object pairs and recognizes the interactions between each pair in an image.  It has been widely applied in scene graph generation \cite{ssg1,ssg2}, visual question answering \cite{vqa1,vqa2}, and action understanding \cite{action1,action2}. Despite significant advancements in recent years, most existing works are restricted to predicting pre-defined HOI categories in the training set. This is mainly because labeling all interacting $<$human, verb, object$>$ triplets in an image is extremely time-consuming, affecting the training set's size and diversity. These facts motivate the study of open vocabulary HOI detection methods. 

Since large Vision-Language Models (VLMs) \cite{clip,blip,blip2} have a strong generalization ability, recent works \cite{gen,eoid,2023hoiclip,clip4hoi,unihoi,synhoi,cmmp} have resorted to VLMs to facilitate open vocabulary HOI detection. They typically rely on the output $cls\_token$ embedding \cite{vit,gen,eoid,synhoi} or the feature maps by the visual encoder of VLMs \cite{2023hoiclip,clip4hoi,unihoi,cmmp}. For example, GEN-VLKT \cite{gen} transfers the knowledge of a VLM to the HOI detector via knowledge distillation according to the $cls\_token$ embedding. MP-HOI \cite{synhoi} fuses the $cls\_token$ embedding with the interaction features extracted by the HOI detector via element-wise addition. Other approaches obtain the knowledge of VLMs using cross-attention \cite{2023hoiclip,clip4hoi,unihoi} or ROI pooling \cite{cmmp,bcom,ezhoi} on the feature maps produced by the visual encoder of VLMs. 

However, the outputs of the VLM visual encoders are coarse-grained \cite{tagclip,clearclip,sclip}. This is due to the training strategy of VLMs, which only involves aligning the holistic features between each image-caption pair, causing the $cls\_token$ embedding to represent holistic information and making the differences in the visual features between image patches ambiguous. This contradicts with the nature of HOI detection, which requires fine-grained instance-level features. Moreover, as illustrated in Figure \ref{fig:intro1}(a), there lacks early interaction between the VLM and HOI detector in existing one-stage methods, which may bring in difficulty in fusing their final output embeddings. 

\begin{figure}[h]
    \centering
    \begin{minipage}[c]{\linewidth}
    \begin{minipage}[c]{0.03\linewidth}
        \centering
        \scriptsize \text{(a)}
    \end{minipage}
    \begin{minipage}[c]{0.97\linewidth}
        \centering
        \includegraphics[width=\linewidth]{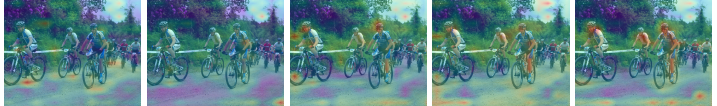} 
    \end{minipage}
    \end{minipage}
    \vfill
    \begin{minipage}[c]{\linewidth}
    \begin{minipage}[c]{0.03\linewidth}
        \centering
        \scriptsize \text{(b)}
    \end{minipage}
    \begin{minipage}[c]{0.97\linewidth}
        \centering
        \includegraphics[width=\linewidth]{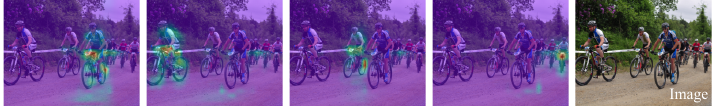}
    \end{minipage}
    \end{minipage}
    \caption{Visualization of attention maps. (a) Self-attention maps for the $cls\_token$ of the ViT encoder of BLIP-2. These maps are extracted from the shallow (left) to deep (right) layers. It is shown that the attention of the $cls\_token$ tends to spread across the entire image. (b) Cross-attention maps for high-score HOI queries in the HOI detector’s interaction decoder. It is shown that each query focuses on local areas of one specific human-object pair.}
    \label{fig:ab1}
\end{figure}

Therefore, we propose a novel one-stage framework that features a bilateral collaboration between the VLM and HOI detector for open vocabulary HOI detection, as illustrated in Figure \ref{fig:intro1}(b). First, we introduce an Attention Bias Guidance (ABG) approach that enables the VLM visual encoder to extract fine-grained features with the guidance from the HOI detector. Specifically, we discover that the cross-attention map for each HOI query in the HOI detector focuses on the interaction area of one specific human-object pair, as shown in Figure \ref{fig:ab1}(b). Hence, we duplicate the $cls\_token$ of the VLM visual encoder to match the number of HOI queries. For each duplicated $cls\_token$, we fuse its attention map in each layer with the cross-attention map of a corresponding HOI query via element-wise addition. With the biased attention maps, we enable VLM to produce a set of $cls\_token$ embeddings that are fine-grained and semantically aligned with the HOI decoder embeddings. Finally, we fuse the two types of embeddings via element-wise addition, ultimately using them for interaction prediction. 

Second, we enhance the power of ABG using a Large Language Model (LLM)-based Supervision Guidance (LSG) approach. LSG improves the ability of the HOI detector to generate high-quality attention bias via the LLM component of the VLM. First, we employ a state-of-the-art Multi-modal LLM (MLLM) to caption all the training images within the HOI detection dataset. Then, we obtain captions that contain rich object- and interaction-relevant descriptions of the image. After that, we add another set of $cls\_token$s to the VLM visual encoder and the same number of extra HOI queries to the HOI detector. Next, we employ the ABG method to extract fine-grained $cls\_token$ embeddings, which are utilized as the input of the LLM component for the VLM. Finally, we use the LLM to predict each token in the above captions. This is a more fine-grained supervision method than common holistic feature-level distillation \cite{gen,eoid,clhoi}. Moreover, we highlight the prediction loss on the nouns and verbs in the captions, which coincides with the concept of HOI detection.

Furthermore, we propose a strategy for fast and parallel computation of ABG. Notably, nearly all VLM model parameters are frozen during training and the LLM component is dropped during inference, significantly enhancing our framework’s efficiency. To the best of our knowledge, our approach is the first to establish close bilateral collaborations between the HOI detector and the VLM for one-stage open vocabulary HOI detection. We demonstrate the effectiveness of our approaches through comprehensive experiments on two popular HOI detection benchmarks: HICO-DET \cite{hico} and V-COCO \cite{unihoi-15}. It is shown that our method consistently achieves state-of-the-art performances in the open vocabulary and closed settings. 

\section{Related Work}
\label{sec:Related Work}
\textbf{HOI Detector Structures.} Existing HOI detection methods can be categorized into one-stage and two-stage paradigms. The two-stage methods usually adopt an off-the-shelf object detector to detect persons and objects in an image. Then, they construct a comprehensive interaction representation for each human-object pair according to multi-stream features such as human and object appearance features \cite{doq31,doq45}, spatial features \cite{vsgnet,scg}, pose features \cite{no,transferable,pose,viplo}, and language features \cite{drg,consnet}. Some works further promote the performance of two-stage models via graph neural networks \cite{qi2018learning,scg,contextual}, which utilize a message-passing mechanism to enhance the human and object features using contextual information. The main disadvantage of two-stage methods is low efficiency, as they enumerate all human-object pairs during the interaction prediction.

In comparison, the one-stage methods search for interactive human-object pairs without the help of object detectors. Initial approaches usually adopt a single point \cite{ppdm,doq30}, a set of points \cite{glance}, or a union region \cite{uniondet} to represent the location of an interactive human-object pair, resulting in a risk of losing a wider range of image context. Recent approaches tend to employ Detection Transformer (DETR) as the backbone. Thanks to the cross-attention operation in DETR, these methods are usually more powerful and more efficient in exploring image-wide context. Moreover, their definitions in the HOI query are flexible. Some methods \cite{qpic,eoid,doq} adopt only a single query to both detect one interactive human-object pair and recognize their interactions. Other methods \cite{cdn,gen,2023hoiclip,unihoi} define multiple types of queries that play different roles. For example, one set of queries is used to detect interactive human-object pairs. Their obtained decoder embeddings are further utilized as the second set of queries for interaction prediction purposes \cite{cdn}. 

In this study, we build our model according to the one-stage paradigm. Unlike existing works, we establish bilateral collaborations between the one-stage HOI detector and a large VLM for open vocabulary HOI detection. 

% Single-query methods use a single query to aggregate all relative information for a human-object pair \cite{qpic,eoid,doq}, while multi-query methods use different queries to aggregate different information essential for interaction prediction, such as information of human, object, and even crucial context \cite{gen,2023hoiclip,synhoi,unihoi}. One-stage detectors are efficient due to their concise frameworks. Our work aims to construct a one-stage detector with robust generalization capabilities.

\textbf{Open Vocabulary HOI Detection.} Since annotating $<$human, verb, object$>$ triplets is time-consuming, the size and diversity of the existing HOI detection databases are limited. Therefore, open vocabulary HOI detection becomes an important research topic in recent years. Existing methods can be grouped into three categories. Methods of the first category rely on compositional learning to promote zero-shot HOI detection \cite{clip4hoi_2,clip4hoi_15,vcl,atl,fcl,dcl,consnet,clip4hoi_36}. Methods of the second category focus on large-scale pre-training \cite{rlip,rlipv2,dphoi}.
For example, Yuan et al. \cite{rlip,rlipv2} obtained pseudo labels for HOI detection from the annotations within external datasets. They further design a data engine that achieves rapid pseudo-label annotation. Li et al. \cite{dphoi} asserted that it is more efficient to use object detection and action recognition datasets to pre-train the detection and interaction decoder layers, respectively, compared to directly using HOI pseudo labels for pre-training the entire model. Methods of the third category are resorted to VLMs \cite{gen,eoid,2023hoiclip,clip4hoi,synhoi,unihoi,cmmp,bcom,ezhoi}. They can be further grouped into the following two types: (1) The methods that employ VLMs in the training stage only. These methods usually transfer the knowledge of the VLM to the HOI detector via knowledge distillation \cite{gen,eoid}; (2) The methods that adopt the VLM in both training and testing stages. These methods usually fuse the output $cls\_token$ embedding or the feature maps of a VLM's visual encoder to improve the interaction prediction accuracy \cite{2023hoiclip,unihoi,clip4hoi,cmmp,synhoi,bcom,ezhoi}. 

In this study, we identify and address two key issues in existing VLM-based approaches: the inadequacy in extracting fine-grained visual features from VLM and the insufficiency in collaboration between the VLM and HOI detector. Finally, we significantly improve the model’s open vocabulary HOI detection performance.

\section{Method}

\definecolor{special_blue}{RGB}{154,198,248}
\definecolor{special_yellow}{RGB}{255,245,200}

\begin{figure*}[htp]
    \centering
    \includegraphics[width=\linewidth]{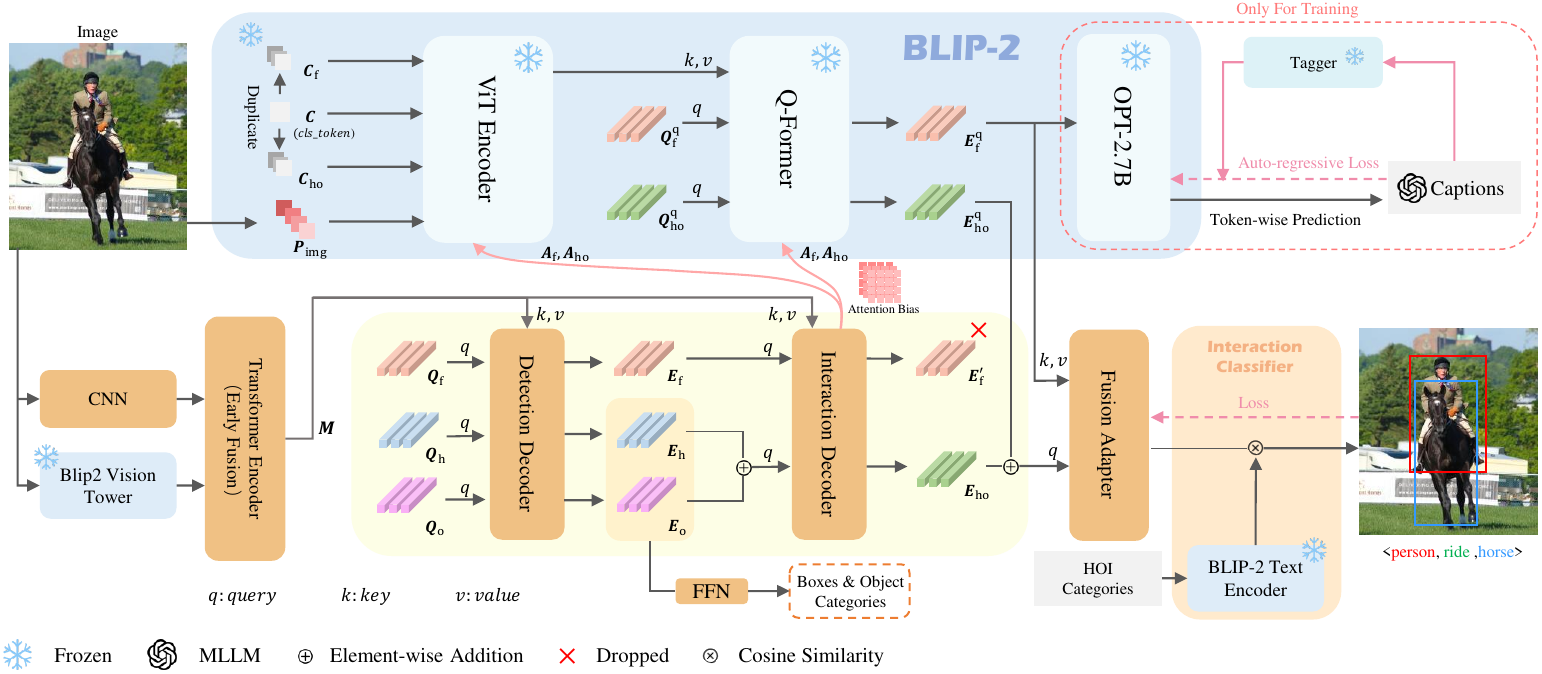}
    \caption{Overview of our bilateral collaboration framework between the HOI detector and VLM. The collaborations are reflected in three aspects: (1) The early fusion between the visual features extracted by the BLIP-2 Vision Tower and the CNN backbone of the HOI detector. (2) The HOI detector provides attention bias to BLIP-2, enabling its Vision Tower to extract fine-grained instance-level interaction features. (3) The BLIP-2 model utilizes its LLM component (OPT-2.7B) to impose fine-grained supervision on the HOI detector. The \textbf{\textcolor{special_blue}{Blue}} shadow represents the BLIP-2 model, while the \textbf{\textcolor{special_yellow}{yellow}} one represents the HOI detector. The LLM is abandoned during the inference stage.}
    \label{fig:method1}
\end{figure*}

In this section, we explain our bilateral collaboration framework between the HOI detector and VLM in detail. We first introduce the overview of our framework in Section \ref{3.1}. Then, we present the guidance from the HOI Detector to VLM in Section \ref{3.2} and the guidance from VLM to the HOI Detector in Section \ref{3.3}, respectively.

\subsection{Overview}
\label{3.1}

Similar to existing works for open vocabulary HOI detection \cite{2023hoiclip,unihoi}, our framework includes one HOI detector and one VLM. We adopt the same HOI detector as \cite{gen}. Specifically, the HOI detector includes one CNN backbone, one transformer encoder \cite{Transformer}, one detection decoder, and one interaction decoder. The detection decoder has two groups of learnable queries as input (i.e., $\bm{Q}_\text{h} \in \mathbb{R}^{N_\text{q}\times D}$ and $\bm{Q}_\text{o} \in \mathbb{R}^{N_\text{q}\times D}$) for human and object detection, respectively, as shown in Figure \ref{fig:method1}. $N_\text{q}$ and $D$ denote the query number and embedding dimension, respectively. The output embeddings of the detection decoder are denoted as $\bm{E}_\text{h} \in \mathbb{R}^{N_\text{q}\times D}$ and $\bm{E}_\text{o} \in \mathbb{R}^{N_\text{q}\times D}$, which are first fused via element-wise addition and then utilized as the queries for the interaction decoder. Moreover, as illustrated in Figure \ref{fig:method1}, we adopt BLIP-2 \cite{blip2} as the VLM following \cite{unihoi}, which includes one ViT encoder \cite{vit}, one Q-Former \cite{blip2}, and one LLM model named OPT-2.7B \cite{opt}. For simplicity, we refer to the cascading ViT encoder and Q-Former as the Vision Tower of BLIP-2, which functions as a visual encoder.

Existing one-stage methods usually adopt late fusion for the features extracted by the HOI detector and VLM \cite{2023hoiclip,unihoi}. Specifically, they employ the interaction decoder and HOI queries to extract representations from feature maps output by the backbone of the HOI detector and the visual encoder of VLM, respectively. The obtained interaction representations are fused via element-wise addition and used for interaction classification purposes, as illustrated in Figure \ref{fig:intro1}(a). 

In comparison, we introduce the early fusion strategy. Specifically, we first concatenate the feature maps produced by the CNN backbone and the BLIP-2 Vision Tower along the spatial dimension. Then, we adopt the transformer encoder to refine the concatenated feature maps. The refined feature map is denoted as $\bm{M} \in \mathbb{R}^{(h\times w + N_\text{p})\times D}$, where the $h$ and $w$ represent the height and width of the CNN feature maps, and $N_\text{p}$ denotes the number of patch tokens in BLIP-2. Compared to the late fusion approach, the early fusion strategy can capture more fine-grained visual feature correlations between the two backbones. Then, both the detection and interaction decoders perform cross-attention on $\bm{M}$.
In the Section \ref{4.4}, we demonstrate that our early fusion strategy improves the HOI detection performance.

\subsection{Guidance from the HOI Detector to VLM}
\label{3.2}
As explained in Section \ref{sec:intro} and illustrated in Figure \ref{fig:ab1}, the features produced by the Vision Tower of BLIP-2 are generalizable but coarse-grained. While the HOI detector is less generalizable, it is better at perceiving fine-grained local features. To combine their advantages, we propose the Attention Bias Guidance (ABG) approach, which guides BLIP-2’s Vision Tower to output instance-level interaction features that are both generalizable and fine-grained. Specifically, ABG approach is based on the cross-attention maps $\bm{A}_\text{ho} \in \mathbb{R}^{N_\text{q}\times h \times w}$ from the interaction decoder:
\begin{equation}
    (\bm{E}_\text{ho}, \bm{A}_\text{ho}) = f^\text{i}(\bm{M},\bm{E}_\text{h} \oplus \bm{E}_\text{o}), \label{eq:1}
\end{equation}
where $f^\text{i}$ denotes the interaction decoder, $\bm{E}_\text{h} \oplus \bm{E}_\text{o}$ is the interaction query for $f^\text{i}$, $\bm{E}_\text{ho} \in \mathbb{R}^{N_\text{q}\times D}$ represents the output embeddings of the interaction decoder, and $\oplus$ stands for element-wise addition. Notably, we only utilize the cross-attention weights relevant to the CNN features in ABG. Therefore, the spatial dimension of $\bm{A}_\text{ho}$ is $h\times w$ instead of $h\times w + N_\text{p}$. Since there are multiple cross-attention layers with each layer containing multiple heads in the interaction decoder, we average all cross-attention maps relevant to a HOI query, obtaining its attention weights in $\bm{A}_\text{ho}$.

\begin{figure}[htp]
    \centering
    \includegraphics[width=\linewidth]{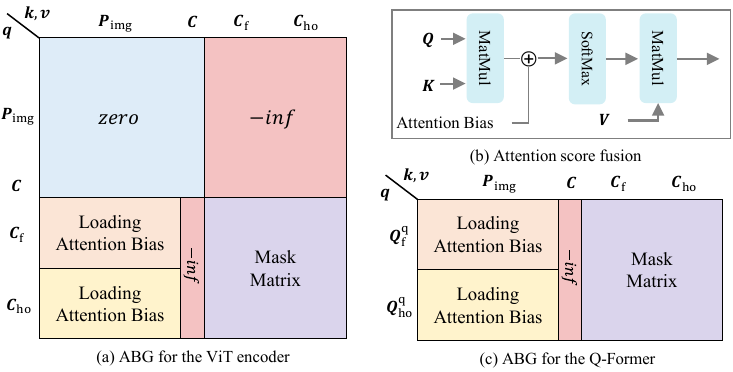}
    \caption{Fast computation of ABG. The $zero$ and -$inf$ matrices denote those with elements that are $0$ and -$inf$, respectively. Diagonal elements in the `mask matrix’ are $0$ and the others are -$inf$.}
    \label{fig:method2}
\end{figure}

In BLIP-2’s ViT encoder, the $cls\_token$ (denoted as $\bm{C} \in \mathbb{R}^{1\times d}$ in Figure \ref{fig:method1}) aggregates the features of all patch tokens $\bm{P}_\text{img} \in \mathbb{R}^{N_\text{p}\times d}$. We duplicate the $cls\_token$ by $N_\text{q}$ folds to obtain $\bm{C}_\text{ho} \in \mathbb{R}^{N_\text{q} \times d}$, so as to match the number of embeddings in $\bm{E}_\text{ho}$. We freeze all embeddings in $\bm{C}_\text{ho}$ and expect each one to focus on the interaction areas of one specified human-object pair. To achieve this, we add each cross-attention map in $\bm{A}_\text{ho}$ as bias to the self-attention maps for the corresponding duplicated $cls\_token$s in $\bm{C}_\text{ho}$, as illustrated in Figure \ref{fig:method2}(b). Specifically, each attention map in $\bm{A}_\text{ho}$ is firstly up-sampled to a fixed size. Then, it is reshaped to a vector and further adapted by a fully-connected (FC) layer. Finally, the processed attention vector is reshaped back to matrix and added to the self-attention maps of BLIP-2. Moreover, we add attention bias in each self-attention layer of BLIP-2’s ViT encoder.

We adopt the similar strategy to the Q-Former part of BLIP-2. Particularly, the Q-Former contains a set of cross-attention layers and originally employs 32 queries that play similar roles to that of $cls\_token$ in the ViT encoder. We propose to enrich the input tokens for the Q-Former with the output embeddings of $\bm{C}$, $\bm{C}_\text{ho}$, and $\bm{P}_\text{img}$ from the ViT encoder. Moreover, we increase the number of queries of Q-Former from 32 to $N_\text{q}$, so as to match the number of HOI queries. This new set of queries is learnable, and  denoted as $\bm{Q}^\text{q}_\text{ho}$ in Figure \ref{fig:method1}. Then, we apply the same attention bias as that in the ViT encoder to ensure the output embeddings of Q-Former are also fine-grained. These $N_\text{q}$ output embeddings are denoted as $\bm{E}^\text{q}_\text{ho}$. The whole process to obtain $\bm{E}^\text{q}_\text{ho}$ can be represented as follows:
\begin{equation}
   \bm{E}^\text{q}_\text{ho} = \phi(\bm{P_\text{img}},\bm{C},\bm{C}_\text{ho};\bm{Q}^\text{q}_\text{ho}|\bm{A}_\text{ho}), \label{eq:2}
\end{equation}
where $\phi$ denotes the Vision Tower, $\bm{C}$ and $\bm{C}_\text{ho}$ are frozen $cls\_token$s, and $\bm{Q}^\text{q}_\text{ho}$ are learnable queries for Q-Former.

Finally, we project $\bm{E}^\text{q}_\text{ho}$ onto the same feature dimension as $\bm{E}_\text{ho}$ using an FC layer and then fuse them via element-wise addition. The fused embeddings can be directly used for interaction classification purposes. 

\textbf{Fast Computation of ABG.} The input tokens for the ViT encoder include $\bm{C}$, $\bm{C}_\text{ho}$, and $\bm{P}_\text{img}$. However, when they play the role of query in the self-attention operation, they have different receptive fields: the receptive fields for $\bm{C}$ and $\bm{P}_\text{img}$ do not include $\bm{C}_\text{ho}$, and the receptive field of each embedding in $\bm{C}_\text{ho}$ only include itself and $\bm{P}_\text{img}$. Moreover, the self-attention maps for $\bm{C}_\text{ho}$ are influenced by the attention bias from the HOI detector, while those for $\bm{C}$ and $\bm{P}_\text{img}$ are irrelevant to the attention bias. These differences impact our model’s computation efficiency. Therefore, we propose unifying the self-attention operations for $\bm{C}$ and $\bm{P}_\text{img}$ into the ABG framework as $\bm{C}_\text{ho}$. 

As illustrated in Figure \ref{fig:method2}(a), we stack the embeddings for $\bm{C}$, $\bm{C}_\text{ho}$ and $\bm{P}_\text{img}$ together to perform the self-attention operations. Then, we impose different attention bias for different query-key pairs. An attention bias of 0 implies that the attention score for this query-key pair remains unchanged. An attention bias of -$inf$ indicates that the query-key pair can be disregarded. In other words, the receptive field of this query does not include this key. When the query is one token from $\bm{C}_\text{ho}$ and the key is one token in $\bm{P}_\text{img}$, we adopt the attention bias from the HOI detector. Finally, the above strategy can be extended to the Q-Former, as illustrated in Figure \ref{fig:method2}(c). Notably, we apply ABG to the cross-attention operations in Q-Former, so the attention maps in Figure \ref{fig:method2}(c) are not square.\\

\subsection{Guidance from VLM to the HOI Detector}
\label{3.3}
The ABG approach relies on the attention maps produced by the HOI detector. In the following, we propose an LLM-based Supervision Guidance (LSG) approach to enhance the quality of these attention maps. This approach works by requiring the LLM component of BLIP-2 to generate more fine-grained and interaction-relevant captions according to the attention bias provided by ABG.

Specifically, as illustrated in Figure \ref{fig:method1}, we add another set of queries $\bm{Q}_\text{f} \in \mathbb{R}^{N_\text{f} \times D}$ to the detection decoder of the HOI detector, a set of duplicated frozen $cls\_token$s $\bm{C}_\text{f} \in \mathbb{R}^{N_\text{f} \times d}$ to the ViT encoder of BLIP-2, and a set of $N_\text{f}$ learnable queries $\bm{Q}^\text{q}_\text{f}$ to the Q-Former of BLIP-2. The output embeddings for $\bm{Q}_\text{f}$ from the detection decoder are denoted as $\bm{E}_\text{f} \in \mathbb{R}^{N_\text{f} \times D}$, which are further utilized as the query for the interaction decoder. Based on the ABG approach, we obtain the cross-attention maps of $\bm{E}_\text{f}$, which are denoted as $\bm{A}_\text{f} \in \mathbb{R}^{N_\text{f}\times h \times w}$. Moreover, with the attention bias contained in $\bm{A}_\text{f}$, we obtain a set of $N_\text{f}$ embeddings $\bm{E}^\text{q}_\text{f}$ from the Q-Former of BLIP-2. Fast computation of ABG in this step is illustrated in Figure \ref{fig:method2}(c).

Then, we feed the embeddings in $\bm{E}^\text{q}_\text{f}$ as tokens to the LLM, prompting it to produce fine-grained captions in the auto-regressive manner \cite{gpt}. In our approach, we obtain the `ground-truth’ captions via a state-of-the-art MLLM (e.g., GPT-4o \cite{gpt4}). Specially, we prompt the MLLM to produce captions that contain object locality information (e.g., “on the left” and “on the bottom”) to match the nature of the detection tasks. More details regarding the captioning process are provided in the supplementary material. Moreover, since the HOI labels contain verbs and nouns only, we apply different weights on the predicted tokens according to the part-of-speech:
\begin{align}
L_\text{lsg} &= \frac{\gamma}{N_\text{t}}{\sum^{N_\text{t}}_{n=1}w_{n}*\text{CELoss}(\bm{t}_n,\bm{g}_n)},\label{eq:3} \\
\bm{t}_n &= f^\text{opt}(\bm{E}^\text{q}_\text{f}|\bm{g}_{1},\bm{g}_{2},...,\bm{g}_{n-1}),
    \label{eq:4}
\end{align}
where $L_\text{lsg}$ denotes our modified auto-regressive loss in BLIP-2, and ${N_\text{t}}$ is the number of valid tokens. In addition, $f^\text{opt}$ represents the LLM component (OPT-2.7B) of BLIP-2. $\gamma$, $w_{n}$, $\bm{t}_n$, and $\bm{g}_n$ denote the loss weight, token weight, predicted logits, and the ground-truth one-hot label vector for the $n$-th token, respectively. Besides, the value of $w_{n}$ for noun, verb, and all the other tokens are set to $\alpha$, $\beta$, and 1.0, respectively.
Since we freeze all model parameters of BLIP-2 except for $\bm{Q}^\text{q}_\text{f}$, the gradients produced by $L_\text{lsg}$ mainly influence the optimization of the HOI detector. Therefore, our LSG approach drives the HOI detector to produce attention maps that cover a broader range of semantic concepts,  allowing the LLM to generate fine-grained captions.

Furthermore, we enhance the quality of the interaction representations (i.e., $\bm{E}_\text{ho} \oplus \bm{E}^\text{q}_\text{ho}$) by applying the cross-attention with $\bm{E}^\text{q}_\text{f}$ using a single transformer layer. We refer to this layer as `Fusion Adapter’ in Figure \ref{fig:method1}. Finally, we adopt the same loss functions for object detection and interaction prediction as those in \cite{gen,unihoi}, and we denote these loss functions as $L_\text{hoi}$. Following existing works, we utilize BLIP-2’s text encoder to construct interaction classifiers \cite{gen}, so as to match the open vocabulary interaction recognition setting. Thus, the total loss function of our model can be represented as follows:
\begin{equation}
   L = L_\text{hoi} + L_\text{lsg}.
\end{equation}
\textbf{Discussion.} Multiple studies also employ extra image captions to enhance the generalization ability of HOI detectors. However, they typically extract HOI triplets as pseudo labels from the captions \cite{rlipv2,dphoi} or impose coarse supervision signals (e.g., knowledge distillation using the holistic textual feature extracted from a text encoder \cite{clhoi}). In comparison, we propose to adopt the LLM and auto-regressive loss to impose token-level fine-grained supervisions and also avoid the complicated pseudo-labeling process.

\section{Experiments}

\subsection{Implementation Details}

We evaluate the performance of our model on two popular benchmarks (i.e., HICO-DET \cite{hico} and V-COCO \cite{unihoi-15}). We adopt the AdamW optimizer \cite{adamw} and conduct experiments with a batch size of 16 on eight 3090 GPUs. We set $N_q$, $N_f$, $\alpha$, $\beta$ and $\gamma$ to 64, 32, 1.5, 2, and 0.1, respectively. The initial learning rate is set to $10^{-4}$ and then multiplied by 0.1 after 40 epochs. We initialize model parameters according to the weights of the DETR model \cite{detr} that is pre-trained on MS-COCO \cite{mscoco}. For the other implementation details, we basically follow those in UniHOI \cite{unihoi} to facilitate fair comparisons with recent works.

\subsection{Comparisons in the Open Vocabulary Setting}
\label{ov_analysis}

We compare the performance of our approach with state-of-the-art methods in four open vocabulary evaluation settings, following GEN-VLKT \cite{gen}. The experimental results are summarized in Table \ref{ov_exp}.

\textbf{Unseen Composition (UC).} In this setting, the same objects and verbs appear in the training and testing stages. However, some verb-object combinations that are unseen in the training stage appear in the testing stage. The UC setting includes two sub-settings (i.e., RF-UC and NF-UC), which indicate that a part of the rare and non-rare verb-object combinations are unseen during training, respectively. In the RF-UC setting, our method outperforms the second-best approach by 11.82\% and 4.50\% mAP on the unseen and seen HOI categories, respectively. In the NF-UC setting, our method outperforms the second-best method by 4.14\% and 4.61\% mAP on the unseen and seen HOI categories, respectively. Our method’s improved performance on the unseen HOI categories is significant. These results indicate that our ABG approach provides accurate attention bias even for unseen verb-object combinations, which enables the VLM to extract generalizable interaction representations and compensates for the lack of training data for these combinations. 

\begin{table}[!hbt]
\caption{Performance comparisons on HICO-DET in the open vocabulary settings. \textbf{Bold} represents the best performance, and \underline{underline} indicates the second-best performance.}
\label{ov_exp}
\centering
\renewcommand{\arraystretch}{1}
\setlength{\tabcolsep}{15pt}
\resizebox{0.42\textwidth}{!}{
\begin{tabular}{c|c|ccc}
\hline
\multicolumn{1}{c}{Method} & \multicolumn{1}{c}{Type} & Unseen & Seen & Full \\ 
\hline
\hline
% VCL \cite{vcl}  & RF-UC & 10.06 & 24.28 & 21.43   \\
% ATL \cite{atl} & RF-UC & 9.18 & 24.67 & 21.57     \\
% FCL \cite{fcl} & RF-UC & 13.16 & 24.23 & 22.01       \\
GEN-VLKT \cite{gen} & RF-UC & 21.36 & 32.91 & 30.56    \\
RLIPv2-ParSeDA \cite{rlipv2} & RF-UC & 21.45 & 35.85 & 32.97 \\
HOICLIP \cite{2023hoiclip}  & RF-UC & 25.53 & 34.85 & 32.99    \\
% MP-HOI \cite{synhoi} & RF-UC & - & - & -  \\
DP-HOI \cite{dphoi} & RF-UC & \underline{30.49} &  \underline{36.17} & \underline{35.03}  \\
UniHOI (w/ BLIP2) \cite{unihoi} & RF-UC & 28.68 & 33.16 & 32.27  \\
Ours (w/ BLIP2)  & RF-UC & \textbf{42.31} & \textbf{40.67} & \textbf{40.99}  \\ 
\hline
\hline
% VCL \cite{vcl} & NF-UC & 16.22 & 18.52 & 18.06       \\
% ATL \cite{atl} & NF-UC & 18.25 & 18.78 & 18.67      \\
% FCL \cite{fcl} & NF-UC & 18.66 & 19.55 & 19.37        \\
GEN-VLKT \cite{gen} & NF-UC & 25.05 & 23.38 & 23.71       \\
RLIPv2-ParSeDA \cite{rlipv2} & NF-UC & 22.81 & 29.52 & 28.18 \\
HOICLIP \cite{2023hoiclip} & NF-UC & 26.39 & 28.10 & 27.75  \\
% MP-HOI \cite{synhoi} & NF-UC & - & - & -  \\
DP-HOI \cite{dphoi} & NF-UC & \underline{28.87} & 29.98 & 29.76  \\
UniHOI (w/ BLIP2) \cite{unihoi} & NF-UC & 28.45 & \underline{32.63} & \underline{31.79}   \\
Ours (w/ BLIP2) & NF-UC & \textbf{33.01} & \textbf{37.24} & \textbf{36.40}   \\
\hline
\hline
GEN-VLKT  \cite{gen} & UO & 10.51 & 28.92 & 25.63          \\
% RLIPv2-ParSeDA \cite{rlipv2} & UO & - & - & - \\
HOICLIP \cite{2023hoiclip} & UO & 16.20 & 30.99 & 28.53  \\
% MP-HOI \cite{synhoi} & UO & - & - & -  \\
% DP-HOI \cite{dphoi} & UO & - & -  &  -  \\
UniHOI (w/ BLIP2) \cite{unihoi} & UO &  \underline{19.72} & \underline{34.76} & \underline{31.56}  \\ 
Ours (w/ BLIP2) & UO & \textbf{19.94} & \textbf{37.03} & \textbf{34.18} \\ 
\hline
\hline
GEN-VLKT \cite{gen} & UV & 20.96 & 30.23 & 28.74    \\
% RLIPv2-ParSeDA \cite{rlipv2} & UV & - & - & - \\
HOICLIP \cite{2023hoiclip} & UV & 24.30 & 32.19 & 31.09  \\
% MP-HOI \cite{synhoi} & UV & - & - & -  \\
DP-HOI \cite{dphoi}& UV & \underline{26.30} & 34.49 & 33.34  \\
UniHOI (w/ BLIP2) \cite{unihoi} & UV & 26.05 & \underline{36.78} & \underline{34.68}  \\
Ours (w/ BLIP2) & UV & \textbf{31.18} & \textbf{41.31} & \textbf{39.89}  \\\hline
\end{tabular}}
\end{table}

\textbf{Unseen Object (UO).} In this setting, some objects are unseen in the training stage but appear during the testing stage. It is shown that our approach achieves the best performance, and outperforms the second-best performance by 0.22\%, 2.27\%, and 2.62\% on Unseen, Seen and Full, respectively. As many object categories are unseen during training in this setting, the ability of the HOI detector to provide attention bias degrades and the performance improvement on Unseen is not significant. We conjecture our performance on the unseen categories can be further improved through pre-training our model with the LSG on large-scale image-text pairs that involve rich object categories.

\textbf{Unseen Verb (UV).} In this setting, some verbs are unseen in the training stage but appear during the testing stage. Our method consistently achieves a significantly higher performance than state-of-the-art methods. It outperforms the second-best method by 4.88\%, 4.53\%, and 5.21\% on the Unseen, Seen, and Full, respectively. This is because compared with existing approaches, our approach is better at extracting fine-grained interaction representations from VLM, which compensates for the lack of verb annotations. 

\begin{table}[!hbt]
\caption{Comparisons on HICO-DET in the closed setting.}
\label{hico_exp}
\centering
\renewcommand{\arraystretch}{1}
\setlength{\tabcolsep}{10pt}
\resizebox{0.42\textwidth}{!}{
\begin{tabular}{c|c|c|ccc}
\hline
\multirow{10.5}*{\rotatebox{90}{Two-Stage}}
&\multicolumn{1}{|c}{}& \multicolumn{1}{c}{}& \multicolumn{3}{c}{Default Setting} 
\\ \cline{4-6} 
&\multicolumn{1}{|c}{Method} &\multicolumn{1}{c}{Backbone} & Full      &   Rare     & Non-Rare     \\ 
\hline
\hline
% &ATL \cite{atl} & ResNet-50 & 23.81& 17.43& 27.42      \\ 
% &VSGNet \cite{vsgnet} & ResNet-152 & 19.80& 16.05& 20.91   \\ 
% &DJ-RN \cite{dj-rn} & ResNet-50 & 21.34& 18.53& 22.18  \\ 
% &VCL \cite{vcl} &  ResNet-50 & 23.63& 17.21& 25.55     \\ 
% &DRG \cite{drg} & ResNet-50 & 24.53& 19.47& 26.04  \\ 
% &IDN \cite{idn} &  ResNet-50 & 24.58& 20.33& 25.86     \\ 
% &FCL \cite{fcl} & ResNet-50 & 25.27 &20.57& 26.67     \\ 
% &SCG \cite{scg} &  ResNet-50 & 29.26& 24.61& 30.65  \\ 
&UPT \cite{upt} & ResNet-50 & 31.66& 25.94& 33.36     \\ 
&CLIP4HOI \cite{clip4hoi} & ResNet-50 & 35.33 & 33.95 & 35.74 \\
&CMMP (w/ ViT-L) \cite{cmmp} & ResNet-50 & 38.14 & 37.75 & 38.25  \\
&ViPLO  \cite{viplo} &  ViT-B/16& 37.22& 35.45& 37.75   \\ 
&EZ-HOI \cite{ezhoi} & ResNet-50 & 38.61  &37.70 & 38.89 \\
&BCOM \cite{bcom} & ResNet-50 &39.34 &39.90 &39.17 \\
\hline
\hline
\multirow{12}*{\rotatebox{90}{One-Stage}}
% &HOI-Trans \cite{hoi-trans} & ResNet-101 & 26.61 & 19.15 & 28.84      \\
% &AS-Net \cite{asnet} & ResNet-50 & 28.87 & 24.25 & 30.25           \\
&QPIC \cite{qpic} & ResNet-101 & 29.90 & 23.92 & 31.69        \\
% &SSRT \cite{ssrt} & ResNet-101 & 31.34 & 24.31 & 33.32           \\
&CDN \cite{cdn} & ResNet-50 & 31.44 & 27.39 & 32.64          \\
&DOQ \cite{doq} & ResNet-50 & 33.28 & 29.19 & 34.50 \\
&GEN-VLKT \cite{gen} & ResNet-50 & 33.75 & 29.25 & 35.10      \\
&RLIPv2-ParSeDA \cite{rlipv2} & ResNet-50 & 35.38 & 29.61 & 37.10  \\ 
&HOICLIP \cite{2023hoiclip} & ResNet-50 & 34.69 & 31.12 & 35.74     \\
&QAHOI \cite{qahoi} & Swin-L & 35.78 & 29.80 & 37.56 \\
&FGAHOI \cite{fgahoi} & Swin-L & 37.18 & 30.71 & 39.11 \\
&MP-HOI \cite{synhoi} & ResNet-50 & 36.50 & 35.48 & 36.80   \\
&DP-HOI\cite{dphoi}& ResNet-50 & 36.56 & 34.36 & 37.22 \\
&UniHOI (w/ BLIP-2) \cite{unihoi} & ResNet-50 & \underline{40.06} & \underline{39.91} & \underline{40.11}  \\
&Ours (w/ BLIP-2) & ResNet-50 & \textbf{43.01} & \textbf{45.76} & \textbf{42.18} \\
% &Ours +  DP-HOI \cite{dphoi} (w/ BLIP-2) & ResNet-50 & \textbf{?} & \textbf{?} & \textbf{?} \\ 
\hline 
\end{tabular}}
\end{table}

\begin{table}[!hbt]
\caption{Comparisons on V-COCO in the closed setting.}
\label{vcoco_exp}
\centering
\renewcommand{\arraystretch}{1}
\setlength{\tabcolsep}{10pt}
\resizebox{0.3\textwidth}{!}{
\begin{tabular}{c|c|ccc}
\hline
\multirow{8.5}*{\rotatebox{90}{Two-Stage}}
&\multicolumn{1}{|c}{Method}& $mAP^{\#1}_{role}$& $mAP^{\#2}_{role}$     \\ 
\hline 
\hline 
% &VSGNet \cite{vsgnet} & 51.8 & 57.0     \\
% &IDN \cite{idn} & 53.3 & 60.3               \\
&UPT \cite{upt} & 59.0 & 64.5              \\
&CLIP4HOI \cite{clip4hoi}  & - & 66.3 \\
&CMMP (w/ ViT-L) \cite{cmmp} & - & 64.0 \\
&VIPLO \cite{viplo} & 62.2 & 68.0             \\  
&EZ-HOI \cite{ezhoi} & 60.5 & 66.2        \\
&BCOM \cite{bcom} & 65.8 & \underline{69.9} \\
\hline
\hline 
\multirow{10}*{\rotatebox{90}{One-Stage}}
% &HOTR \cite{hotr} &  55.2 & 64.4                  \\
&QPIC \cite{qpic} &  58.8 & 61.0                  \\
&CDN \cite{cdn} &  61.68 & 63.77              \\
&GEN-VLKT \cite{gen} & 62.41 & 64.46               \\
&RLIPv2-ParSeDA \cite{rlipv2} & 65.9 & 68.0 \\
&HOICLIP \cite{2023hoiclip}&  63.50 & 64.80  \\
&FGAHOI (Swin-T) \cite{fgahoi} & 60.5 & 61.2 \\
&MP-HOI \cite{synhoi} & 66.2 & 67.6 \\
&DP-HOI \cite{dphoi} & \underline{66.6} & - \\
&UniHOI (w/ BLIP2) \cite{unihoi} & 65.58 & 68.27   \\
&Ours (w/ BLIP2) & \textbf{68.20} & \textbf{70.61} \\ \hline
\end{tabular}}
\end{table}

\subsection{Comparisons in the Closed Setting}
\label{hico_analysis}

\textbf{Performance on HICO-DET.} As shown in Table \ref{hico_exp}, our approach significantly outperforms all the one- and two-stage methods, achieving state-of-the-art results. Specifically, our method outperforms the second-best approach (i.e., UniHOI \cite{unihoi}), by 2.95\%, 5.85\%, and 2.07\% on Full, Rare, and Non-rare, respectively.  Notably, except for the Early Fusion (EF), ABG, and LSG strategies, we adopt very similar model structure and training schemes as UniHOI \cite{unihoi}. Moreover, the performance of our model on the rare HOI categories is superior. The above experimental results indicate that our ABG and LSG strategies are successful in extracting generalizable and instance-level interaction representations from the VLM. This reduces the impact of the lack of training data on the rare HOI category predictions. \\
\textbf{Performance on V-COCO.} The performance comparison results on V-COCO are in Table \ref{vcoco_exp}. Our approach outperforms all the other methods, achieving 68.20\% in terms of $mAP^{\#1}_{role}$ and 70.61\% in terms of $mAP^{\#2}_{role}$. \\
The above comparisons demonstrate our approach achieves superior performance in both closed and open vocabulary settings, indicating its practicality in a broad range of real-world applications. We further compare the model complexity between our method and UniHOI \cite{unihoi} in Table \ref{efficiency}. During the training phase, we utilize 8 NVIDIA 3090 GPUs with a batch size of 2. UniHOI requires training for 90 epochs, and our model needs 60 epochs. During the inference stage, we use a single NVIDIA 3090 GPU with a batch size of 1. It is shown that our model is smaller during inference. Considering the significant performance improvements, the slight additional time cost of single-image inference and the complete training is acceptable.

\begin{table}[!hbt]
\caption{Comparisons in training and inference cost.}
\label{efficiency}
\centering
\renewcommand{\arraystretch}{1}
\setlength{\tabcolsep}{5pt}
\resizebox{0.45\textwidth}{!}{
\begin{tabular}{c|ccc|ccc}
\hline
\multicolumn{1}{c}{Method} & \multicolumn{3}{c}{Inference} & \multicolumn{3}{c}{Training} \\ 
\hline
 &Time(per image) & Model size & GPU & Time(per epoch) & Model size & GPU \\ 
\hline
UniHOI\cite{unihoi} & 82.3ms & 1140.05M & 11.99G & 13min18s all = 19.95h & 1140.05M  & 13.74G\\
BC-HOI & 99.7ms & 1138.62M & 11.97G & 18min57s all = 22.10h & 3790.22M  & 16.68G\\
\hline
\end{tabular}}
\end{table}

\begin{table}[!hbt]
\caption{Ablation studies on the closed setting of HICO-DET. `EF’ denotes early fusion.}
\label{close_ab}
\centering
\renewcommand{\arraystretch}{1}
\setlength{\tabcolsep}{15pt}
\resizebox{0.4\textwidth}{!}{
\begin{tabular}{c|ccc|ccc}
\hline
\multicolumn{1}{c}{} &  &  &\multicolumn{1}{c}{} &\multicolumn{3}{c}{Default Setting} \\ \cline{5-7} 
\multicolumn{1}{c}{Methods} & EF & ABG & \multicolumn{1}{c}{LSG}& Full      & Rare     & Non-rare     \\ 
\hline
\hline
Baseline& - & - & - & 38.91 & 36.27 & 39.70     \\
\hline
\hline
\multirow{4}*{Ours}
& \checkmark & - & -  & 39.34 & 37.35 & 39.94        \\
& \checkmark & \checkmark & -  & 42.60 & 44.87 & 41.92        \\
& \checkmark & - & \checkmark  & 42.41 & 43.71 & 42.03     \\
% & \checkmark & \textbullet & \textbullet &41.22 
%  & 41.42  & 41.15     \\
& \checkmark & \checkmark & \checkmark  & \textbf{43.01} & \textbf{45.76} & \textbf{42.18}  \\
\hline
\end{tabular}}
\end{table}

\begin{table}[!hbt]
\caption{Ablation studies on the NF-UC setting of HICO-DET.}
\label{ov_ab}
\centering
\renewcommand{\arraystretch}{1}
\setlength{\tabcolsep}{11pt}
\resizebox{0.4\textwidth}{!}{
\begin{tabular}{c|ccc|ccc}
\hline
\multicolumn{1}{c}{Methods} & EF & ABG & \multicolumn{1}{c}{LSG}& Unseen & Seen & Full     \\ 
\hline
\hline
Baseline& - & - & - & 29.24 & 31.71 & 31.22     \\
\hline
\hline
\multirow{4}*{Ours}
& \checkmark & - & -  & 29.97 & 32.17 & 31.73      \\
& \checkmark & \checkmark & -  & 31.70 & 35.97 & 35.12       \\
& \checkmark & - & \checkmark  & 31.37 & 33.84 & 33.38     \\
% & \checkmark & \textbullet & \textbullet  & 30.52 & 33.14 & 32.60     \\
& \checkmark & \checkmark & \checkmark  & \textbf{33.01} & \textbf{37.24} & \textbf{36.40}  \\
\hline
\end{tabular}}
\end{table}

\begin{table*}[!hbt]
\caption{Comparisons with variants of our approach on the NF-UC setting of HICO-DET.}
\label{variants}
\centering
\begin{tabular}{@{}ccc@{}}
\begin{minipage}[t]{0.3\textwidth}
{\small{(a) Attention Bias Sources}}
\centering
\renewcommand{\arraystretch}{1}
\setlength{\tabcolsep}{8pt}
\resizebox{\textwidth}{!}{
\begin{tabular}{c|ccc}
\hline
\multicolumn{1}{c}{Method} & Unseen & Seen & Full \\ 
\hline
\hline
Detection Decoder Only & 32.55 & 36.15 & 35.43 \\
Interaction Decoder Only & \textbf{33.01} & \textbf{37.24} & \textbf{36.40} \\
Both & 32.76 & 36.34 & 35.62 \\
\hline
\end{tabular}}
\end{minipage}
&
\begin{minipage}[t]{0.3\textwidth}
{\small (b) Attention Bias Destinations}
\centering
\renewcommand{\arraystretch}{1}
\setlength{\tabcolsep}{12pt}
\resizebox{\textwidth}{!}{
\begin{tabular}{c|ccc}
\hline
\multicolumn{1}{c}{Method} & Unseen & Seen & Full \\ 
\hline
\hline
ViT encoder Only & 32.48 & 36.19 & 35.45 \\
Q-Former Only & 32.15 & 36.03 & 35.25 \\
Both & \textbf{33.01} & \textbf{37.24} & \textbf{36.40} \\
\hline
\end{tabular}}
\end{minipage}
\begin{minipage}[t]{0.3\textwidth}
{\small (c) Fixed vs Adaptive Attention Bias}
\centering
\renewcommand{\arraystretch}{1}
\setlength{\tabcolsep}{10pt}
\resizebox{\textwidth}{!}{
\begin{tabular}{c|ccc}
\hline
\multicolumn{1}{c}{Method} & Unseen & Seen & Full \\ 
\hline
\hline
Plain Learnable Bias & 30.52 & 33.14 & 32.60 \\
Ours & \textbf{33.01} & \textbf{37.24} & \textbf{36.40} \\
\hline
\end{tabular}}
\end{minipage}
\end{tabular}
\\
\begin{tabular}{@{}ccc@{}}
\begin{minipage}[t]{0.29\textwidth}
{\small {(d) Variants of LSG Supervision}}
\centering
\renewcommand{\arraystretch}{1}
\setlength{\tabcolsep}{10pt}
\resizebox{\textwidth}{!}{
\begin{tabular}{c|ccc}
\hline
\multicolumn{1}{c}{Method} & Unseen & Seen & Full \\ 
\hline
\hline
Caption Level & 32.13 & 36.24 & 35.56 \\
w/o $Q^\text{q}_\text{f}$ & 32.56 & 36.55 & 35.75 \\
% w/o Fusion Adapter & ? & ? & ? \\
% w/o Special Weights & 32.57 & 36.64 & 35.82 \\
Ours & \textbf{33.01} & \textbf{37.24} & \textbf{36.40} \\
\hline
\end{tabular}}
\end{minipage}
&
\begin{minipage}[t]{0.28\textwidth}
{\small(e) Sources of Captions}
\centering
\renewcommand{\arraystretch}{1}
\setlength{\tabcolsep}{10pt}
\resizebox{\textwidth}{!}{
\begin{tabular}{c|ccc}
\hline
\multicolumn{1}{c}{Method} & Unseen & Seen & Full \\ 
\hline
\hline
w/o LSG  & 31.70 & 35.97 & 35.12 \\
QwenVL-7B & 32.76 & 37.11 & 36.24 \\
GPT-4o & \textbf{33.01} & \textbf{37.24} & \textbf{36.40}  \\
\hline
\end{tabular}}
\end{minipage}
&
\begin{minipage}[t]{0.29\textwidth}
{\small {(f) Value of $w_\text{n}$}}
\centering
\renewcommand{\arraystretch}{1}
\setlength{\tabcolsep}{8.5pt}
\resizebox{\textwidth}{!}{
\begin{tabular}{ccc|ccc}
\hline
\multicolumn{1}{c}{$\alpha$}& \multicolumn{1}{c}{$\beta$} & \multicolumn{1}{c}{$else$} & Unseen & Seen & Full \\ 
\hline
\hline
1 & 1 & 0 & 32.47 & 36.56 & 35.74 \\
1 & 1 & 1 & 32.79 & 36.98 & 36.14 \\
1.5 & 2 & 1 & \textbf{33.01} & \textbf{37.24} & \textbf{36.40} \\
\hline
\end{tabular}}
\end{minipage}
\end{tabular}
\end{table*}

\subsection{Ablation Studies}
\label{4.4}

In this subsection, we conduct experiments to validate the effectiveness of each component. The experiments are conducted in the NF-UC and closed settings of the HICO-DET database. To construct a baseline model, we remove the EF, ABG, and LSG components from our approach. It adopts the Late Fusion (LF) strategy only \cite{2023hoiclip,unihoi} to fuse the interaction features extracted from the HOI detector and original visual features produced by VLM. More baseline details are provided in the supplementary. We conduct experiments using different components on the baseline model. The results are displayed in Tables \ref{close_ab}, \ref{ov_ab}, and \ref{variants}.

\textbf{Effectiveness of EF.} In Tables \ref{close_ab} and \ref{ov_ab}, we add the EF component to the baseline model. It is shown that EF promotes HOI detection in both settings. These results justify that EF effectively captures the correlations between the feature maps produced by the two backbones.

\textbf{Effectiveness of ABG.} In this experiment, we add both EF and ABG components to the baseline model. Then, we conduct another complementary experiment where we keep the EF and LSG components but remove ABG. The results in Tables \ref{close_ab} and \ref{ov_ab} show that compared to the experiments with EF only, EF+ABG promotes HOI detection performance by 3.26\% (Full) and 1.73\% (Unseen) in closed setting and NF-UC setting, respectively. When ABG is removed from our whole framework, the performance declines by 0.60\% (Full) and 1.64\% (Unseen). Notably, in the closed setting, a significant performance boost occurs on Rare when ABG component is enabled. In the NF-UC setting, this phenomenon also occurs on seen categories. These results validate that ABG effectively enables VLM to extract generalized and instance-level interaction features. 

\textbf{Effectiveness of LSG.} In this experiment, we conduct two ablation experiments similar to ABG. The results in Tables \ref{close_ab} and \ref{ov_ab} show that when LSG is added to EF, there are notable performance changes. In the closed setting, the performance improves by 3.07\%, 6.36\%, and 2.09\% on Full, Rare, and Non-Rare, respectively. In the NF-UC setting, performance improves by 1.40\%, 1.67\%, and 1.65\% on Unseen, Seen, and Full, respectively. These results indicate that the supervision from LSG is effective.

\textbf{Comparisons with variants of ABG.} In Table \ref{variants}(a), we compare the performance of our model with different attention bias sources. Three strategies are considered: attention maps from the detection decoder only, attention maps from the interaction decoder only, and both. In the third situation, we average the attention maps from the detection and interaction decoders. It is shown that the second strategy achieves the best performance. This is reasonable since it is the interaction features to extract from the VLM.

In Table \ref{variants}(b), we compare the performance of our model with different ABG destinations. There are also three strategies considered: ViT encoder only, Q-Former only, and both. It is shown that each of three strategies promotes performance and the third one performs the best. 

Finally, in Table \ref{variants}(c), we replace the attention bias produced by ABG with plain learnable parameters. The experimental results are significantly lower than those of ABG. This is because the plain learnable parameters are fixed during inference. In comparison, the attention bias provided by ABG are adaptive to each image and enables $\bm{E}^\text{f}_\text{ho}$ to be semantically aligned with $\bm{E}_\text{ho}$. 

\textbf{Comparisons with variants of LSG.} In Table \ref{variants}(d), we first replace our token-level LSG supervision with the coarse caption-level supervision. In this experiment, we feed the caption into the text encoder of BLIP-2, and then employ the obtained holistic textual feature to supervise the learning of $E^\text{q}_\text{f}$. Specifically, we average the embeddings in $E^\text{q}_\text{f}$ and penalize the cosine distance between this averaged embedding and the holistic textual feature. It is shown that this variant achieves lower performance than ours. This is because holistic textual features mainly contain coarse information, while HOI detection is a fine-grained task. We also conduct another experiment, as shown in Table \ref{variants}(d), where we remove $Q^\text{q}_\text{f}$. Instead, we employ $E^\text{q}_\text{ho}$ as the input tokens of the LLM. The result is lower than that of our design, which indicates HOI detection and caption generation are different tasks. Therefore, the LLM component requires different inputs to better accomplish these two tasks. 

Moreover, we compare the performance of LSG with the `ground-truth captions’ generated by the QwenVL-7B \cite{qwen} and GPT-4o \cite{gpt4} models, respectively. The experimental results are summarized in Table \ref{variants}(e). These results show that LSG consistently improves model performance with each of two types of captions. 
Finally, we test different token weights in Eq.\ref{eq:3} for the predicted tokens according to their part-of-speech. The experimental results are displayed in Table \ref{variants}(f) and the supplementary material. It is shown that predicting all words is more efficient than predicting the verbs and nouns alone, and imposing larger weights on the verbs and nouns can further improve model performance. The former maybe because useful image context and locality information (e.g., `on the left') can be considered by predicting all the words in the caption. The latter maybe due to the verbs and nouns being more essential for the HOI detection task than the other words.

\section{Conclusions and Limitations}

In this study, we propose a bilateral collaboration framework for one-stage open vocabulary HOI detection, which includes three key components, namely Early Fusion, Attention Bias Guidance (ABG), and LLM-based Supervision Guidance (LSG). In ABG, we utilize attention bias produced by the HOI detector to guide the VLM to output both generalizable and instance-level interaction features. In LSG, we utilize the LLM component of VLM to provide token-level fine-grained supervisions for the HOI detector, which further promotes the power of ABG. We conduct extensive experiments on popular benchmarks and consistently achieve superior performance on all settings. Our work also has certain limitations. For example, the introduction of VLM results in higher GPU memory cost.\\
\textbf{Broader Impacts.} Our model detects all human-object pairs of interest in an image and infers the interactions between each pair. It has a wide range of applications, e.g., robotics and smart home systems. To the best of our knowledge, it does not have obvious negative social impacts. \\
\textbf{Acknowledgement.} This work was supported by the National Natural Science Foundation of China under Grant 62476099 and 62076101, Guangdong Basic and Applied Basic Research Foundation under Grant 2024B1515020082 and 2023A1515010007, the Guangdong Provincial Key Laboratory of Human Digital Twin under Grant 2022B1212010004, the TCL Young Scholars Program, and the 2024 Tencent AI Lab Rhino-Bird Focused Research Program.

% {
%     \small
%     \bibliographystyle{ieeenat_fullname}
%     \bibliography{ref}

\clearpage
\begin{thebibliography}{26}\itemsep=-1pt

\bibitem{2023hoiclip}\label{1}
S. Ning, L. Qiu, Y. Liu, X. He. Hoiclip: Efficient knowledge transfer for hoi detection with vision-language models. In {\it{CVPR}}, 2023.

\bibitem{clip4hoi}\label{2}
Y. Mao, J. Deng, W. Zhou, L. Li, Y. Fang, H. Li. CLIP4HOI: towards adapting CLIP for practical zero-shot HOI detection. In {\it{NeurIPS}}, 2023.

\bibitem{horcnn}\label{3}
Y. Chao, Y. Liu, X. Liu, H. Zeng, J. Deng. Learning to detect human-object interactions. In {\it{WACV}}, 2018.

\bibitem{gen}\label{4}
Y. Liao, A. Zhang, M. Lu, Y. Wang, X. Li, S. Liu. Gen-vlkt: Simplify association and enhance interaction understanding for hoi detection. In {\it{CVPR}}, 2022.

\bibitem{vcl}\label{5}
Z. Hou, X. Peng, Y. Qiao, D. Tao. Visual compositional learning for human-object interaction detection. In {\it{ECCV}}, 2020.

\bibitem{eoid}\label{6}
M. Wu, J. Gu, Y. Shen, M. Lin, C. Chen, X. Sun. End-to-end zero-shot hoi detection via vision and language knowledge distillation. In {\it{AAAI}}, 2023.

\bibitem{drg}\label{7}
C. Gao, J. Xu, Y. Zou, J. Huang. Drg: Dual relation graph for human-object interaction detection. In {\it{ECCV}}, 2020.

\bibitem{asnet}\label{8}
M. Chen, Y. Liao, S. Liu, Z. Chen, F. Wang, C. Qian. Reformulating hoi detection as adaptive set prediction. In {\it{CVPR}}, 2021.

\bibitem{qpic}\label{9}
M. Tamura, H. Ohashi, T. Yoshinaga. Qpic: Query-based pairwise human-object interaction detection with image-wide contextual information. In {\it{CVPR}}, 2021.

\bibitem{upt}\label{10}
F. Zhang, D. Campbell, S. Gould. Efficient two-stage detection of human-object interactions with a novel unary-pairwise transformer. In {\it{CVPR}}, 2022.

\bibitem{doq}\label{11}
X. Qu, C. Ding, X. Li, X. Zhong, D. Tao. Distillation using oracle queries for transformer-based human-object interaction detection. In {\it{CVPR}}, 2022.

\bibitem{cdn}\label{12}
A. Zhang, Y. Liao, S. Liu, M. Lu, Y. Wang, C. Gao, X. Li. Mining the benefits of two-stage and one-stage hoi detection. In {\it{NeurIPS}}, 2021.

\bibitem{cql}\label{13}
C. Xie, F. Zeng, Y. Hu, S. Liang, Y. Wei. Category query learning for human-object interaction classification. In {\it{CVPR}}, 2023.

\bibitem{unihoi}\label{14}
Y. Cao, Q. Tang, X. Su, S. Chen, S. You, X. Lu, C. Xu. Detecting any human-object interaction relationship: Universal hoi detector with spatial prompt learning on foundation models. In {\it{NeurIPS}}, 2023.

\bibitem{cmmp}\label{15}
T. Lei, S. Yin, Y. Peng, Y. Liu. Exploring conditional multi-modal prompts for zero-shot hoi detection. In {\it{ECCV}}, 2024.

\bibitem{synhoi}\label{16}
J. Yang, B. Li, F. Yang, A. Zeng, L. Zhang, R. Zhang. Open-World Human-Object Interaction Detection via Multi-modal Prompts. In {\it{CVPR}}, 2024.

\bibitem{unihoi-15}\label{17}
S. Gupta, J. Malik. Visual semantic role labeling. In arXiv:1505.04474, 2015.

\bibitem{vsgnet}\label{18}
O. Ulutan, A. Iftekhar, B. Manjunath. Vsgnet: Spatial attention network for detecting human object interactions using graph convolutions. In {\it{CVPR}}, 2020.

\bibitem{adacm}\label{19}
T. Lei, F. Caba, Q. Chen, H. Jin, Y. Peng, Y. Liu. Efficient Adaptive Human-Object Interaction Detection with Concept-guided Memory. In {\it{ICCV}}, 2023.

\bibitem{no}\label{20}
T. Gupta, A. Schwing, D. Hoiem. No-frills human-object interaction detection: Factorization, layout encodings, and training techniques. In {\it{ICCV}}, 2019.

\bibitem{transferable}\label{21}
Y. Li, S. Zhou, X. Huang, L. Xu, Z. Ma, H. Fang, Y. Wang, C. Lu. Transferable interactiveness knowledge for human-object interaction detection. In {\it{CVPR}}, 2019.

\bibitem{pose}\label{22}
B. Wan, D. Zhou, Y. Liu, R. Li, X. He. Pose-aware multi-level feature network for human object interaction detection. In {\it{ICCV}}, 2019.

\bibitem{viplo}\label{23}
J. Park, J. Park, J. Lee. Viplo: Vision transformer based pose-conditioned self-loop graph for human-object interaction detection. In {\it{CVPR}}, 2023.

\bibitem{consnet}\label{24}
Y. Liu, J. Yuan, C. Chen. Consnet: Learning consistency graph for zero-shot human-object interaction detection. In {\it{ACM MM}}, 2020.

\bibitem{qi2018learning}\label{25}
S. Qi, W. Wang, B. Jia, J. Shen, S. Zhu. Learning human-object interactions by graph parsing neural networks. In {\it{ECCV}}, 2018.

\bibitem{contextual}\label{26}
H. Wang, W. Zheng, L. Yingbiao. Contextual heterogeneous graph network for human-object interaction detection. In {\it{ECCV}}, 2020.

\bibitem{ppdm}\label{27}
Y. Liao, S. Liu, F. Wang, Y. Chen, C. Qian, J. Feng. Ppdm: Parallel point detection and matching for real-time human-object interaction detection. In {\it{CVPR}}, 2020.

\bibitem{doq30}\label{28}
T. Wang, T. Yang, M. Danelljan, F. Khan, X. Zhang, J. Sun. Learning human-object interaction detection using interaction points. In {\it{CVPR}}, 2020.

\bibitem{glance}\label{29}
X. Zhong, X. Qu, C. Ding, D. Tao. Glance and gaze: Inferring action-aware points for one-stage human-object interaction detection. In {\it{CVPR}}, 2021.

\bibitem{uniondet}\label{30}
B. Kim, T. Choi, J. Kang, H. Kim. Uniondet: Union-level detector towards real-time human-object interaction detection. In {\it{ECCV}}, 2020.

\bibitem{clip}\label{31}
A. Radford, J. Kim, C. Hallacy, A. Ramesh, G. Goh, S. Agarwal, G. Sastry, A. Askell, P. Mishkin, J. Clark, Learning transferable visual models from natural language supervision. In {\it{ICML}}, 2021.

\bibitem{vild}\label{32}
X. Gu, T. Lin, W. Kuo, Y. Cui. Open-vocabulary object detection via vision and language knowledge distillation. In {\it{ICLR}}, 2022.

\bibitem{glip}\label{33}
L. Li, P. Zhang, H. Zhang, J. Yang, C. Li, Y. Zhong, L. Wang, L. Yuan, L. Zhang, J. Hwang, Grounded language-image pre-training. In {\it{CVPR}}, 2022.

\bibitem{rlip}\label{34}
H. Yuan, J. Jiang, S. Albanie, T. Feng, Z. Huang, D. Ni, M. Tang. Rlip: Relational language-image pre-training for human-object interaction detection. In {\it{NeurIPS}}, 2022.

\bibitem{blip}\label{35}
J. Li, D. Li, C. Xiong, S. Hoi. Blip: Bootstrapping language-image pre-training for unified vision-language understanding and generation. In {\it{ICML}}, 2022.

\bibitem{blip2}\label{36}
J. Li, D. Li, S. Savarese, S. Hoi. Blip-2: Bootstrapping language-image pre-training with frozen image encoders and large language models. In {\it{ICML}}, 2023.

\bibitem{regionclip}\label{37}
Y. Zhong, J. Yang, P. Zhang, C. Li, N. Codella, L. Li, L. Zhou, X. Dai, L. Yuan, Y. Li, Regionclip: Region-based language-image pretraining. In {\it{CVPR}}, 2022.

\bibitem{tagclip}\label{38}
Y. Lin, M. Chen, K. Zhang, H. Li, M. Li, Z. Yang, D. Lv, B. Lin, H. Liu, D. Cai. TagCLIP: A Local-to-Global Framework to Enhance Open-Vocabulary Multi-Label Classification of CLIP without Training. In {\it{AAAI}}, 2024.

\bibitem{cleanclip}\label{39}
H. Bansal, N. Singhi, Y. Yang, F. Yin, A. Grover, K. Chang. Cleanclip: Mitigating data poisoning attacks in multimodal contrastive learning. In {\it{ICCV}}, 2023.

\bibitem{denseclip}\label{40}
Y. Rao, W. Zhao, G. Chen, Y. Tang, Z. Zhu, G. Huang, J. Zhou, J. Lu. Denseclip: Language-guided dense prediction with context-aware prompting. In {\it{CVPR}}, 2022.

\bibitem{rlipv2}\label{41}
H. Yuan, S. Zhang, X. Wang, S. Albanie, Y. Pan, T. Feng, J. Jiang, D. Ni, Y. Zhang, D. Zhao. Rlipv2: Fast scaling of relational language-image pre-training. In {\it{ICCV}}, 2023.

\bibitem{vqa1}\label{42}
Z. Khan, Y. Fu. Consistency and Uncertainty: Identifying Unreliable Responses From Black-Box Vision-Language Models for Selective Visual Question Answering. In {\it{CVPR}}, 2024.

\bibitem{vqa2}\label{43}
L. Li, J. Peng, H. Chen, C. Gao, X. Yang. How to configure good in-context sequence for visual question answering. In {\it{CVPR}}, 2024.

\bibitem{ssg1}\label{44}
R. Li, S. Zhang, D. Lin, K. Chen, X. He. From Pixels to Graphs: Open-Vocabulary Scene Graph Generation with Vision-Language Models. In {\it{CVPR}}, 2024.

\bibitem{ssg2}\label{45}
C. Zhang, S. Stepputtis, J. Campbell, K. Sycara, Y. Xie. HiKER-SGG: Hierarchical Knowledge Enhanced Robust Scene Graph Generation. In {\it{CVPR}}, 2024.

\bibitem{action1}\label{46}
H. Kim, J. Hong, H. Kong, S. Lee. TE-TAD: Towards Full End-to-End Temporal Action Detection via Time-Aligned Coordinate Expression. In {\it{CVPR}}, 2024.

\bibitem{action2}\label{47}
B. Xiong, X. Yang, Y. Song, Y. Wang, C. Xu. Modality-Collaborative Test-Time Adaptation for Action Recognition. In {\it{CVPR}}, 2024.

\bibitem{detr}\label{48}
N. Carion, F. Massa, G. Synnaeve, N. Usunier, A. Kirillov, S. Zagoruyko. End-to-end object detection with transformers. In {\it{ECCV}}, 2020.

\bibitem{hico}\label{49}
Y. Chao, Y. Liu, X. Liu, H. Zeng, J. Deng. A Benchmark for Recognizing Human-Object Interactions in Images. In {\it{ICCV}}, 2015.

\bibitem{resnet}\label{50}
K. He, X. Zhang, S. Ren, J. Sun. Deep residual learning for image recognition. In {\it{CVPR}}, 2016.

\bibitem{adamw}\label{51}
I. Loshchilov. Decoupled weight decay regularization. In {\it{ICLR}}, 2019.

\bibitem{dphoi}\label{52}
Z. Li, X. Li, C. Ding, X. Xu. Disentangled Pre-training for Human-Object Interaction Detection. In {\it{CVPR}}, 2024.

\bibitem{clip4hoi_2}\label{53}
A. Bansal, S. Rambhatla, A. Shrivastava, R. Chellappa. Detecting human-object interactions via functional generalization. In {\it{AAAI}}, 2020.

\bibitem{clip4hoi_15}\label{54}
T. Gupta, A. Schwing, D. Hoiem. No-frills human-object interaction detection: Factorization, layout encodings, and training techniques. In {\it{ICCV}}, 2019.

\bibitem{atl}\label{55}
Z. Hou, B. Yu, Y. Qiao, X. Peng, D. Tao. Affordance transfer learning for human-object interaction detection. In {\it{CVPR}}, 2021.

\bibitem{fcl}\label{56}
Z. Hou, B. Yu, Y. Qiao, X. Peng, D. Tao. Detecting human-object interaction via fabricated compositional learning. In {\it{CVPR}}, 2021.

\bibitem{dcl}\label{57}
Z. Hou, B. Yu, D. Tao. Discovering human-object interaction concepts via self-compositional learning. In {\it{ECCV}}, 2022.

\bibitem{clip4hoi_36}\label{58}
J. Peyre, I. Laptev, C. Schmid, J. Sivic. Detecting unseen visual relations using analogies. In {\it{ICCV}}, 2019.

\bibitem{clearclip}\label{59}
M. Lan, C. Chen, Y. Ke, X. Wang, L. Feng, W. Zhang. Clearclip: Decomposing clip representations for dense vision-language inference. In {\it{ECCV}}, 2024.

\bibitem{sclip}\label{60}
F. Wang, J. Mei, A. Yuille. Sclip: Rethinking self-attention for dense vision-language inference. In {\it{ECCV}}, 2024.

\bibitem{dj-rn}\label{61}
Y. Li, X. Liu, H. Lu, S. Wang, J. Liu, J. Li, C. Lu. Detailed 2d-3d joint representation for human-object interaction. In {\it{CVPR}}, 2020.

\bibitem{idn}\label{62}
Y. Li, X. Liu, X. Wu, Y. Li, C. Lu. Hoi analysis: Integrating and decomposing human-object interaction. In {\it{NeurIPS}}, 2020.

\bibitem{scg}\label{63}
F. Zhang, D. Campbell, S. Gould. Spatially conditioned graphs for detecting human-object interactions. In {\it{ICCV}}, 2021.

\bibitem{hoi-trans}\label{64}
C. Zou, B. Wang, Y. Hu, J. Liu, Q. Wu, Y. Zhao, B. Li, C. Zhang, C. Zhang, Y. Wei, End-to-end human object interaction detection with hoi transformer. In {\it{CVPR}}, 2021.

\bibitem{ssrt}\label{65}
A. Iftekhar, H. Chen, K. Kundu, X. Li, J. Tighe, D. Modolo. What to look at and where: Semantic and spatial refined transformer for detecting human-object interactions. In {\it{CVPR}}, 2022.

\bibitem{qahoi}\label{66}
J. Chen, K. Yanai. Qahoi: Query-based anchors for human-object interaction detection. In {\it{MVA}}, 2023.

\bibitem{fgahoi}\label{67}
S. Ma, Y. Wang, S. Wang, Y. Wei. Fgahoi: Fine-grained anchors for human-object interaction detection. In {\it{TPAMI}}, 2023.

\bibitem{hotr}\label{68}
B. Kim, J. Lee, J. Kang, E. Kim, H. Kim. Hotr: End-to-end human-object interaction detection with transformers. In {\it{CVPR}}, 2021.

\bibitem{doq31}\label{69}
T. Zhou, W. Wang, S. Qi, H. Ling, J. Shen. Cascaded human-object interaction recognition. In {\it{CVPR}}, 2020.

\bibitem{doq45}\label{70}
Y. Liu, Q. Chen, A. Zisserman. Amplifying key cues for human-object-interaction detection. In {\it{ECCV}}, 2020.

\bibitem{doq19}\label{73}
Q. Dong, Z. Tu, H. Liao, Y. Zhang, V. Mahadevan, S. Soatto. Visual relationship detection using part-and-sum transformers with composite queries. In {\it{ICCV}}, 2021.

\bibitem{doq24}\label{74}
M. Chen, Y. Liao, S. Liu, Z. Chen, F. Wang, C. Qian. Reformulating hoi detection as adaptive set prediction. In {\it{CVPR}}, 2021.

\bibitem{vit}\label{75}
A. Dosovitskiy, L. Beyer, A. Kolesnikov, D. Weissenborn, X. Zhai, T. Unterthiner, M. Dehghani, M. Minderer, G. Heigold, S. Gelly, J. Uszkoreit, N. Houlsby. An image is worth 16x16 words: Transformers for image recognition at scale. In {\it{ICLR}}, 2021.

\bibitem{groundingdino}\label{76}
S. Liu, Z. Zeng, T. Ren, F. Li, H. Zhang, J. Yang, C. Li, J. Yang, H. Su, J. Zhu, Grounding dino: Marrying dino with grounded pre-training for open-set object detection. In {\it{ECCV}}, 2024.

\bibitem{opt}\label{77}
S. Zhang, S. Roller, N. Goyal, M. Artetxe, M. Chen, S. Chen, C. Dewan, M. Diab, X. Li, X. Lin, Opt: Open pre-trained transformer language models. In arXiv:2205.01068, 2022.

\bibitem{gpt}\label{78}
A. Radford. Improving language understanding by generative pre-training. In \url{https://openai.com/blog/chatgpt/}, 2018.

\bibitem{gpt4}\label{79}
A. Radford. Improving language understanding by generative pre-training. In \url{https://openai.com/blog/gpt-4/}, 2023.

\bibitem{clhoi}\label{80}
J. Gao, C. Cai, R. Wang, W. Liu, K. Yap, K. Garg, B. Han. CL-HOI: Cross-Level Human-Object Interaction Distillation from Vision Large Language Models. In arXiv:2410.15657, 2024.

\bibitem{mscoco}\label{81}
T. Lin, M. Maire, S. Belongie, J. Hays, P. Perona, D. Ramanan, Microsoft coco: Common objects in context. In {\it{ECCV}}, 2014.

\bibitem{qwen}\label{82}
J. Bai, S. Bai, S. Yang, S. Wang, S. Tan, P. Wang, J. Lin, C. Zhou, J. Zhou. Qwen-vl: A frontier large vision-language model with versatile abilities. In arXiv:2308.12966, 2023.

\bibitem{Transformer}\label{83}
A. Vaswani, N. Shazeer, N. Parmar, J. Uszkoreit, L. Jones, A. Gomez, L. Kaiser, I. Polosukhin. Attention is all you need. In {\it{NeurIPS}}, 2017.

\bibitem{bcom}\label{84}
G. Wang, Y. Guo, Z. Xu, M. Kankanhalli. Bilateral adaptation for human-object interaction detection with occlusion-robustness. In {\it{CVPR}}, 2024.

\bibitem{DIFFUSIONHOI}\label{85}
L. Li, W. Wang, Y. Yang. Human-object interaction detection collaborated with large relation-driven diffusion models. In {\it{NeurIPS}}, 2024.

\bibitem{ezhoi}\label{86}
Q. Lei, B. Wang, R. Tan. EZ-HOI: VLM Adaptation via Guided Prompt Learning for Zero-Shot HOI Detection. In {\it{NeurIPS}}, 2024.

\bibitem{zeng2022graph}\label{67}
R. Zeng, W. Huang, M. Tan, Y. Rong, P. Zhao, J. Huang, C. Gan. Graph Convolutional Module for Temporal Action Localization in Videos.
In {\it{TPAMI}}, 2022.


\end{thebibliography}
% }

\clearpage
\setcounter{page}{1}
\maketitlesupplementary
\appendix

\renewcommand{\thetable}{\Alph{table}}
\setcounter{table}{0}

\renewcommand{\thefigure}{\Alph{figure}}
\setcounter{figure}{0}

This supplementary material includes five sections. Section \ref{A} provides a more comprehensive performance comparison in the closed setting. Section \ref{B} presents our prompt for MLLM to generate captions. Section \ref{C} shows the visualization of attention maps. Section \ref{D} offers a more thorough ablation study on the token weights. More implementation details of the baseline and our model (BC-HOI) are presented in Section \ref{E} and Section \ref{F}, respectively. Section \ref{G} shows the implementation details of “EF Only" and “EF+LSG" settings in the ablation studies, while Section \ref{H} demonstrates the implementation of token-level supervision. Additional ablation studies for “EF+LSG" are presented in Section \ref{I}.

\section{More Performance Comparisons}
\label{A}
In this section, we compare our method with more existing approaches on HICO-DET \cite{hico}, as shown in Table \ref{hico_exp_ko}. To better demonstrate the effectiveness of our method, we provide additional experimental results in the Known-Object (KO) setting. It is shown that our method achieves state-of-the-art performance in both the Default and KO settings. Specifically, in the KO setting, our method outperforms the second-best approach by 3.15\%, 5.34\%, and 2.49\% in the Full, Seen, and Unseen categories, respectively. These experimental results fully demonstrate our method’s strong modeling capability for both rare and non-rare HOI categories. The analysis of experimental results on the Default setting is already presented in Section \ref{hico_analysis}. Similarly, we provide the performance comparison with more existing methods on V-COCO \cite{unihoi-15}, as shown in Table \ref{vcoco_exp_else}.

\begin{table*}[!hbt]
\caption{Performance comparisons on HICO-DET in the Default setting and Known-Object (KO) setting. \textbf{Bold} represents the best performance, and \underline{underline} indicates the second-best performance.}
\label{hico_exp_ko}
\centering
\renewcommand{\arraystretch}{1}
\setlength{\tabcolsep}{10pt}
\resizebox{0.85\textwidth}{!}{
\begin{tabular}{c|c|c|ccc||ccc}
\hline
\multirow{16}*{\rotatebox{90}{Two-Stage}}
&\multicolumn{1}{|c}{}& \multicolumn{1}{c}{}& \multicolumn{3}{c}{Default Setting} & \multicolumn{3}{c}{Known Object}
\\ \cline{4-9} 
&\multicolumn{1}{|c}{Method} &\multicolumn{1}{c}{Backbone} & Full      &   Rare     & Non-Rare & Full & Rare  & Non-Rare    \\ 
\hline
\hline
& ATL \cite{atl} & ResNet-50 & 23.81 & 17.43 & 25.72 & 27.38 & 22.09 & 28.96 \\
&VSGNet \cite{vsgnet} & ResNet-152 & 19.80 & 16.05 & 20.91 & - & - & - \\
&DJ-RN \cite{dj-rn} & ResNet-50 & 21.34 & 18.53 & 22.18 & 23.69 & 20.64 & 24.60 \\
&VCL \cite{vcl} & ResNet-101 & 23.63 & 17.21 & 25.55 & 25.98 & 19.12 & 28.03 \\
&DRG \cite{drg} & ResNet-50-FPN & 24.53 & 19.47 & 26.04 & 27.98 & 23.11 & 29.43 \\
&IDN \cite{idn} & ResNet-50 & 26.29 & 22.61 & 27.39 & 28.24 & 24.47 & 29.37 \\
&UPT \cite{upt} & ResNet-50 & 31.66 & 25.94& 33.36  & 35.05 & 29.27 & 36.77  \\ 
&CLIP4HOI \cite{clip4hoi} & ResNet-50 & 35.33 & 33.95 & 35.74 & 37.19 & 35.27 & 37.77 \\
&CMMP (w/ ViT-L) \cite{cmmp} & ResNet-50 & 38.14 & 37.75 & 38.25 & - & - & - \\
&ViPLO  \cite{viplo} &  ViT-B/16& 37.22& 35.45& 37.75 & 40.61 & 38.82 & 41.15  \\  
&EZ-HOI \cite{ezhoi} & ResNet-50 & 38.61  &37.70 & 38.89 & - & - & -\\
&BCOM \cite{bcom} & ResNet-50 &39.34 &39.90 &39.17 & \underline{42.24} & \underline{42.86} & 42.05 \\
\hline
\hline
\multirow{15.5}*{\rotatebox{90}{One-Stage}}
&PPDM \cite{ppdm} & Hourglass-104 & 21.73 & 13.78 & 24.10 & 24.58 & 16.65 & 26.84 \\
&HOI-Trans \cite{hoi-trans} & ResNet-101 & 26.61 & 19.15 & 28.84 & 29.13 & 20.98 & 31.57 \\
&AS-Net \cite{asnet} & ResNet-50 & 28.87 & 24.25 & 30.25 & 31.74 & 27.07 & 33.14 \\
&QPIC \cite{qpic} & ResNet-101 & 29.90 & 23.92 & 31.69 & 32.38 & 26.06 & 34.27       \\
&CDN \cite{cdn} & ResNet-50 & 31.44 & 27.39 & 32.64 & 34.09 & 29.63 & 35.42         \\
&DOQ \cite{doq} & ResNet-50 & 33.28 & 29.19 & 34.50 & - & - & - \\
&GEN-VLKT \cite{gen} & ResNet-50 & 33.75 & 29.25 & 35.10 & 36.78 & 32.75 & 37.99     \\
&MP-HOI \cite{synhoi} & ResNet-50 & 36.50 & 35.48 & 36.80 & - & - & -   \\
&HOICLIP \cite{2023hoiclip} & ResNet-50 & 34.69 & 31.12 & 35.74 & 37.61 & 34.47 & 38.54     \\
&RLIPv2-ParSeDA \cite{rlipv2} & ResNet-50 & 35.38 & 29.61 & 37.10 & - & - & -   \\ 
&QAHOI \cite{qahoi} & Swin-L & 35.78 & 29.80 & 37.56 & 37.59 & 31.66 & 39.36 \\
&DP-HOI\cite{dphoi}& ResNet-50 & 36.56 & 34.36 & 37.22 & 39.37 & 36.59 & 40.20 \\
&FGAHOI \cite{fgahoi} & Swin-L & 37.18 & 30.71 & 39.11 & 38.93 & 31.93 & 41.02 \\
&UniHOI (w/ BLIP-2) \cite{unihoi} & ResNet-50 & \underline{40.06} & \underline{39.91} & \underline{40.11} & 42.20 & 42.60 &  \underline{42.08}  \\
&Ours (w/ BLIP-2) & ResNet-50 & \textbf{43.01} & \textbf{45.76} & \textbf{42.18} & \textbf{45.35} & \textbf{47.94} & \textbf{44.57} \\
% &Ours +  DP-HOI \cite{dphoi} (w/ BLIP-2) & ResNet-50 & \textbf{?} & \textbf{?} & \textbf{?} \\ 
\hline 
\end{tabular}}
\end{table*}

\section{Prompts of MLLM for Captioning}
\label{B}
\begin{figure}[!t]
    \centering
    \includegraphics[width=0.9\linewidth]{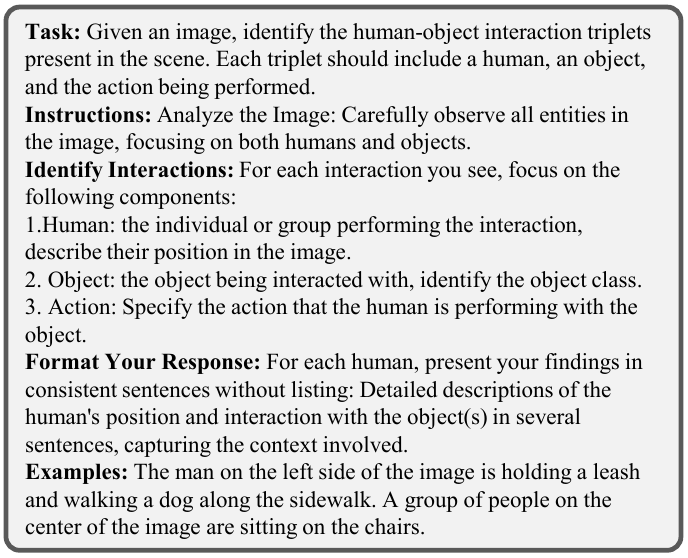}
    \caption{The prompt we adopted to instruct the MLLM to caption images in the LSG component.}
    \label{fig:caption}
\end{figure}

In this section, we demonstrate how we prompt MLLM to generate captions with rich image context and locality information. Specifically, we first provide GPT-4o \cite{gpt4} with a detailed description of the HOI task to obtain a basic prompt, which is then manually refined. Second, we enhance the prompt to enable MLLM to generate captions with locality information using the modified prompt. Additionally, we offer an example in the prompt so as to explain how to describe multi-person scenarios (e.g., using ‘a group of’). The final prompt is illustrated in Figure \ref{fig:caption}.

\begin{table}[!hbt]
\caption{Comparisons on V-COCO in the closed setting.}
\label{vcoco_exp_else}
\centering
\renewcommand{\arraystretch}{1}
\setlength{\tabcolsep}{8pt}
\resizebox{0.35\textwidth}{!}{
\begin{tabular}{c|c|ccc}
\hline
\multirow{12}*{\rotatebox{90}{Two-Stage}}
&\multicolumn{1}{|c}{Method}& $mAP^{\#1}_{role}$& $mAP^{\#2}_{role}$     \\ 
\hline 
\hline 
&VCL \cite{vcl} & 48.3 & - \\
&DRG \cite{drg} & 51.0 & -  \\
&VSGNet \cite{vsgnet} & 51.8 & 57.0     \\
&IDN \cite{idn} & 53.3 & 60.3               \\
&UPT \cite{upt} & 59.0 & 64.5              \\
&CLIP4HOI \cite{clip4hoi}  & - & 66.3 \\
&CMMP (w/ ViT-L) \cite{cmmp} & - & 64.0 \\
&VIPLO \cite{viplo} & 62.2 & 68.0             \\  
&EZ-HOI \cite{ezhoi} & 60.5 & 66.2        \\
&BCOM \cite{bcom} & 65.8 & \underline{69.9} \\
\hline
\hline 
\multirow{13.5}*{\rotatebox{90}{One-Stage}}
&HOI-Trans \cite{hoi-trans} & 52.9 & - \\
&AS-Net \cite{asnet} &53.9 & - \\
&HOTR \cite{hotr} &  55.2 & 64.4                  \\
&QPIC \cite{qpic} &  58.8 & 61.0                  \\
&CDN \cite{cdn} &  61.68 & 63.77              \\
&GEN-VLKT \cite{gen} & 62.41 & 64.46               \\
&RLIPv2-ParSeDA \cite{rlipv2} & 65.9 & 68.0 \\
&HOICLIP \cite{2023hoiclip}&  63.50 & 64.80  \\
&FGAHOI (Swin-T) \cite{fgahoi} & 60.5 & 61.2 \\
&MP-HOI \cite{synhoi} & 66.2 & 67.6 \\
&DP-HOI \cite{dphoi} & \underline{66.6} & - \\
&UniHOI (w/ BLIP2) \cite{unihoi} & 65.58 & \underline{68.27}   \\
&Ours (w/ BLIP2) & \textbf{68.20} & \textbf{70.61} \\ \hline
\end{tabular}}
\end{table}

\section{Visualization of Attention Maps}
\label{C}
\begin{figure*}[!t]
    \centering
    \includegraphics[width=0.9\linewidth]{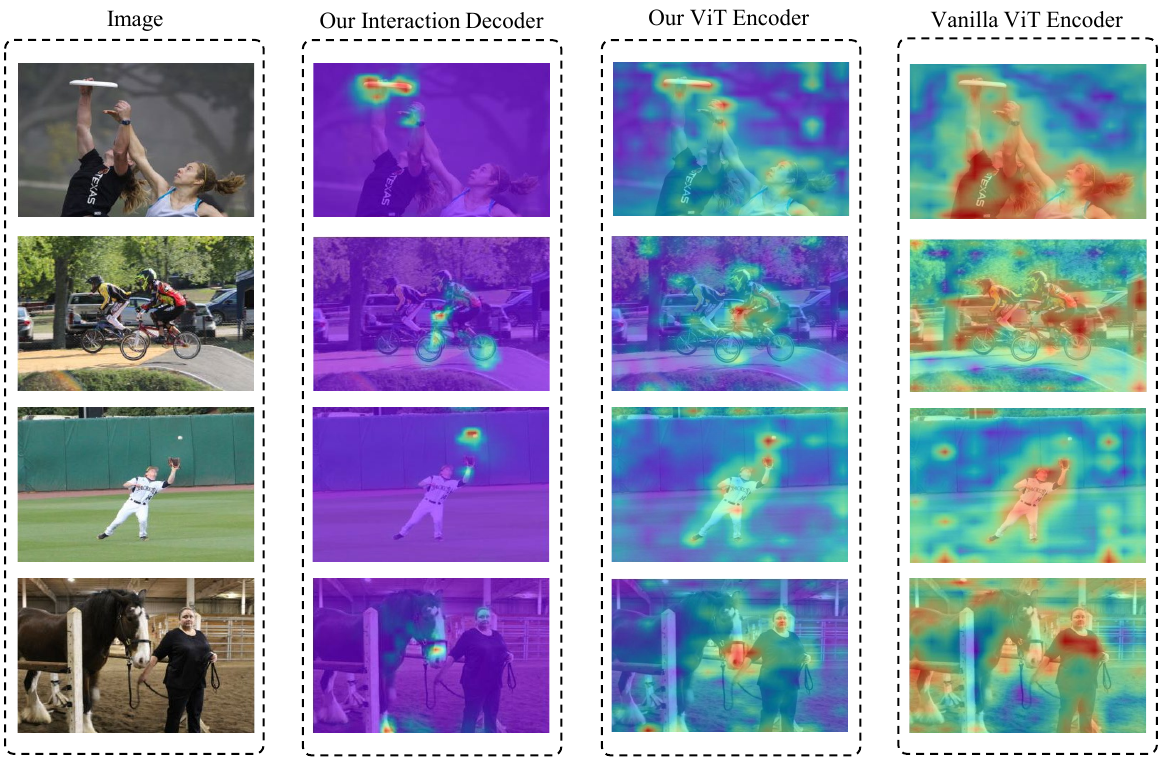}
    \caption{Visualization of the attention maps. Each row demonstrates the original image, the cross-attention map for one selected HOI query in our interaction decoder, the self-attention map of the corresponding $cls\_token$ embedding in $\bm{C}_\text{ho}$, and the self-attention map of the $cls\_token$ embedding in the vanilla ViT encoder of BLIP-2. It is shown that with the guidance by ABG, the self-attention map produced by the $cls\_token$ embedding in $\bm{C}_\text{ho}$ effectively focuses on the interaction areas of an interested human-object pair. In contrast, the self-attention map produced by the vanilla ViT encoder of BLIP-2 attend to all foreground objects in the image.}
    \label{fig:ap1}
\end{figure*}

To validate the effectiveness of our proposed method, we visualize the attention maps from our interaction decoder, our ViT encoder, and the vanilla ViT encoder of BLIP-2, as shown in Figure \ref{fig:ap1}. Specifically, we train our model in the Default setting of HICO-DET \cite{hico}, and extract the cross-attention map for one selected HOI query from the interaction decoder. Then, we obtain the self-attention map of the corresponding $cls\_token$ embedding in $\bm{C}_\text{ho}$, and the self-attention map of the $cls\_token$ embedding in the vanilla ViT encoder of BLIP-2. It is shown that the HOI query and the $cls\_token$ embedding in $\bm{C}_\text{ho}$ can effectively focus on the interaction area of a specific human-object pair in the image. In comparison, the $cls\_token$ embedding in the vanilla ViT encoder of BLIP-2 attend to all the foreground objects in the image. This demonstrates that our HOI detector provides high-quality attention bias via ABG, resulting in a significant HOI detection performance improvement.

\section{Study on the Token Weights}
\label{D}
In this section, we further conduct more experiments on the token weights $w_n$ in Eq.\ref{eq:3} of LSG. Experiments are conducted on the NF-UC setting of the HICO-DET database. The experimental results are presented in Tables \ref{ab_w}. It is shown that the combination of $\alpha = 1.5$, $\beta = 2$, and $else = 1$ performs the best. We assert that it is necessary to impose a different loss weight to each token according to its part-of-speech, so that the model can focus more on HOIs contained in the caption.

\begin{table}
\caption{Study on the value of token weight $w_n$}
\label{ab_w}
\centering
\renewcommand{\arraystretch}{1}
\setlength{\tabcolsep}{5pt}
\resizebox{0.3\textwidth}{!}{
\begin{tabular}{ccc|ccc}
\hline
\multicolumn{1}{c}{$\alpha$}& \multicolumn{1}{c}{$\beta$} & \multicolumn{1}{c}{$else$} & Unseen & Seen & Full \\ 
\hline
\hline
1 & 1 & 0 & 32.47 & 36.56 & 35.74 \\
1 & 1 & 1 & 32.79 & 36.98 & 36.14 \\
2 & 2 & 1 & 32.89 & 37.20 & 36.37 \\
2 & 1.5 & 1 & 32.82 & 37.07 & 36.25 \\
1.5 & 2 & 1 & \textbf{33.01} & \textbf{37.24} & \textbf{36.40} \\
\hline
\end{tabular}}
\end{table}

\section{More Details of Baseline}
\label{E}
\begin{figure*}[htp]
    \centering
    \includegraphics[width=\linewidth]{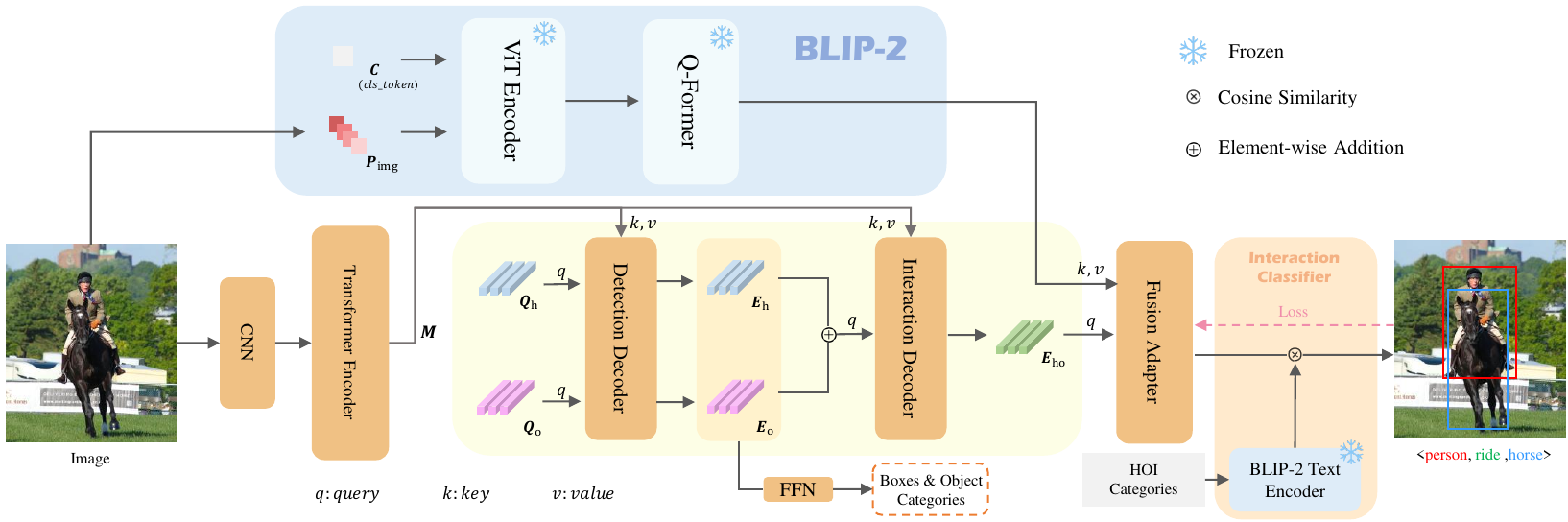}
    \caption{Model structure of the baseline model.}
    \label{fig:baseline}
\end{figure*}

We illustrate the structure of the baseline model in Figure \ref{fig:baseline}. To construct the baseline model, we remove the EF, ABG, and LSG components from BC-HOI. The same as BC-HOI, the baseline adopts one transformer layer to fuse the visual features produced by BLIP-2’s Vision Tower and the interaction features produced by the HOI detector. Specifically, it adopts $\bm{E}_\text{ho}$ as the query, while the output of Q-Former serve as the key and value, respectively. Besides, the baseline adopts the same classifier as BC-HOI. 

\begin{figure*}[htp]
    \centering
    \includegraphics[width=0.9\linewidth]{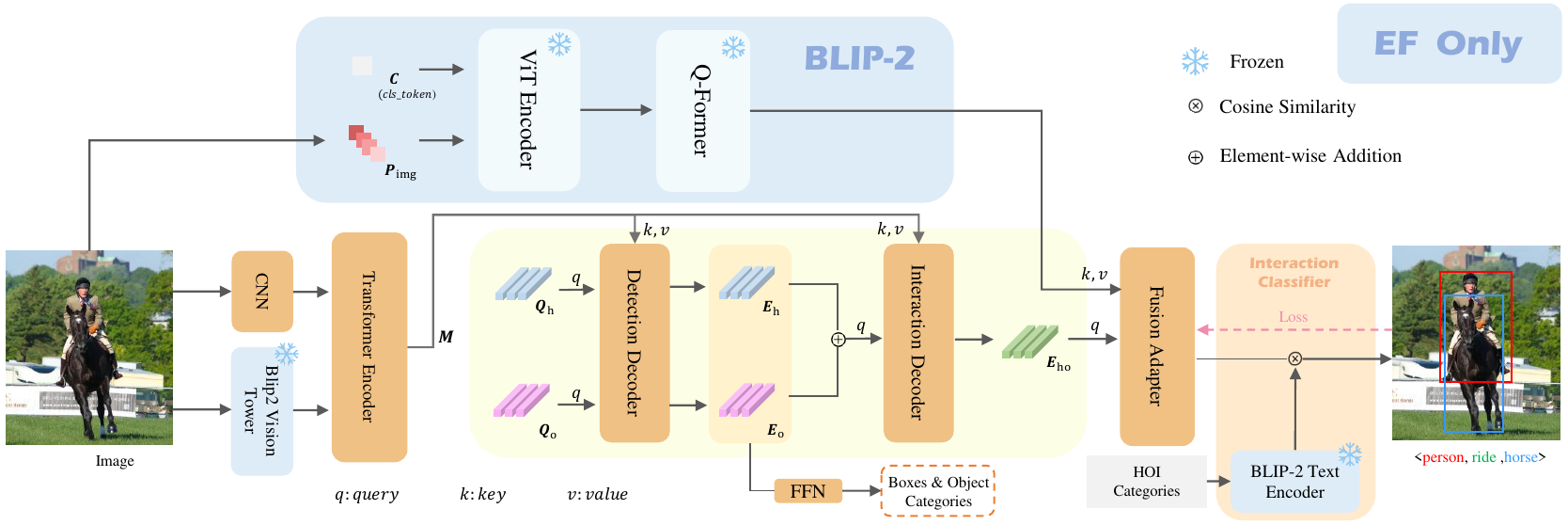}
    \caption{Model structure of the “EF Only" setting in the ablation study.}
    \label{fig:EF}
\end{figure*}

\begin{figure*}[htp]
    \centering
    \includegraphics[width=0.9\linewidth]{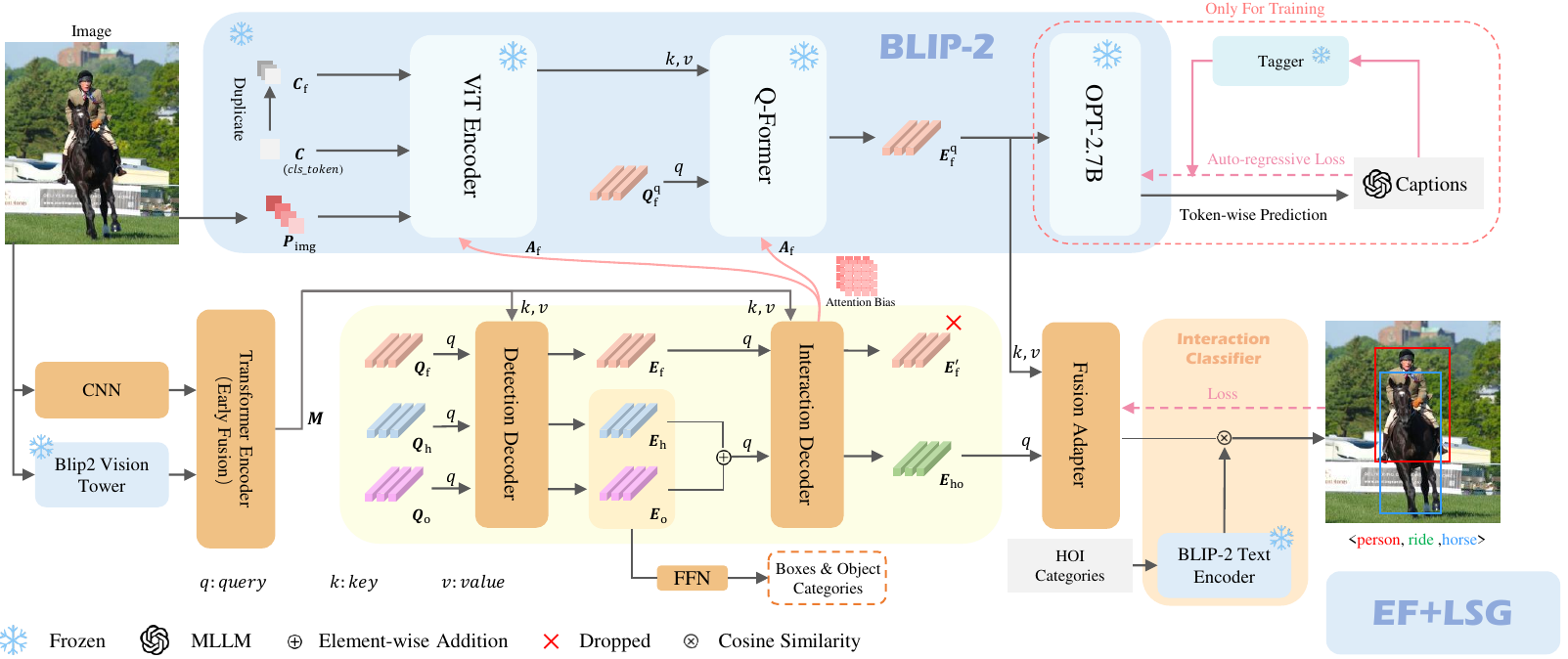}
    \caption{Model structure of the “EF+LSG" setting in the ablation study.}
    \label{fig:EF+LSG}
\end{figure*}

\section{More Details of BC-HOI}
\label{F}
In BC-HOI, the BLIP-2 OPT-2.7B \cite{blip2} is adopted as the large VLM for all experiments following UniHOI \cite{unihoi}. Specifically, each image is first down-sampled to $224 \times 224$ pixels and then being divided into $16 \times 16$ patches. These patches are concatenated with $\bm{C}$, $\bm{C_\text{f}}$, and $\bm{C_\text{ho}}$ as the input of the ViT encoder, producing embeddings with the dimension of $(1 + 16 \times 16 + 64 + 32) \times 1408$. The output embeddings from the ViT encoder are then fed into the Q-Former as keys and values of the cross-attention layers, while $\bm{Q}^\text{q}_\text{ho}$ and $\bm{Q}^\text{q}_\text{f}$ serve as the queries with dimensions of $64 \times 768$ and $32 \times 768$, respectively. The output features $\bm{E}^\text{q}_\text{ho}$ from the Q-Former are fused with the features extracted by the HOI detector via element-wise addition. Another set of output features $\bm{E}^\text{q}_\text{f}$ from the Q-Former is both fed into the OPT-2.7B model for image captioning during training and into the Fusion Adapter as keys and values. Moreover, we adopt “a photo of a person $<$doing something$>$" as template to construct a phrase for each HOI category. This phrase is then fed into the BLIP-2 ViT-L text encoder to obtain the interaction classifier.

\section{More Details of “EF Only" and “EF+LSG"}
\label{G}
For clarity, we illustrate the structure of “EF Only” and “EF+LSG" of the ablation studies in Figures \ref{fig:EF} and \ref{fig:EF+LSG}, respectively. In the “EF Only” setting, we add EF (Early Fusion) to the baseline, which means the branch of BLIP-2 Vision Tower remains in the model. In the “EF+LSG" setting, we remove $\bm{C}_\text{ho}$, $\bm{Q}^\text{q}_\text{ho}$, $\bm{A}_\text{ho}$, and $\bm{E}^\text{q}_\text{ho}$ from Figure 3. Although all $cls\_tokens$ ($\bm{C}$ and $\bm{C}_\text{f}$) attend to the same features ($\bm{P}_\text{img}$), their attention maps are affected by different bias values within $\bm{A}_\text{f}$.

\begin{figure*}[htp]
    \centering
    \includegraphics[width=0.9\linewidth]{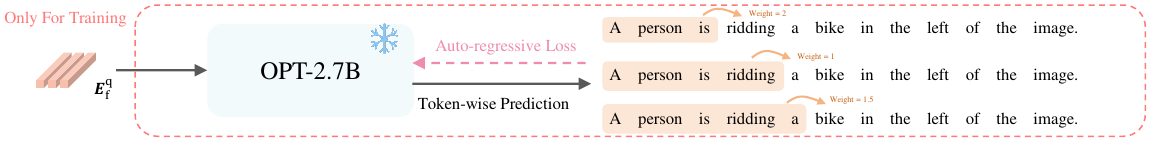}
    \caption{Implementation of the token-level supervision.}
    \label{fig:Loss}
\end{figure*}

\section{More Details of token-level supervision.}
\label{H}
As illustrated in Figure \ref{fig:Loss}, in the LSG framework, we feed the embeddings $\bm{E}^\text{q}_\text{f}$ into BLIP-2's LLM component, prompting it to generate fine-grained captions in an auto-regressive manner. This produces token-level supervision, enabling the model to produce high-quality attention maps. Notably, the code of the auto-regressive loss is highly mature, and computations for each token are processed in parallel, ensuring sufficient efficiency.

\begin{table}
\caption{Study on “EF+LSG" setting in the NF-UC setting}
\label{ab_ef_lsg}
\centering
\renewcommand{\arraystretch}{1}
\setlength{\tabcolsep}{5pt}
\resizebox{0.45\textwidth}{!}{
\begin{tabular}{c|ccc}
\hline
\multicolumn{1}{c}{method} & Unseen & Seen & Full \\ 
\hline
\hline
EF+LSG w/o ($\bm{A}_\text{f}$ and LLM) & 30.38 & 32.71 & 32.10 \\
EF+LSG w/o $\bm{A}_\text{f}$ & 30.62 & 33.10 & 32.57 \\
EF+LSG &31.37 & 33.84 & 33.38 \\
\hline
\end{tabular}}
\end{table} 

\section{Study on “EF+LSG" setting}
\label{I}
To demonstrate the novelty and effectiveness of our LSG method, we provide additional ablation studies in the “EF+LSG" setting, as shown in the Table \ref{ab_ef_lsg}. In the “EF+LSG w/o $\bm{A}_\text{f}$” setting, we remove the influence of $\bm{A}_\text{f}$, and the model performance drops to 30.62\%/33.10\%/32.57\% on Unseen/Seen/Full in the UC-NF setting.
This demonstrates that LSG enhances the performance by optimizing the attention maps generated by the model. In the “EF+LSG w/o ($\bm{A}_\text{f}$ and LLM)" setting, we remove both $\bm{A}_\text{f}$ and LLM from “EF+LSG" and retain the learnable $\bm{Q}^\text{q}_\text{f}$. Compared with the performance of “EF+LSG", the performance drops by 0.99\%/1.13\%/1.28\% on Unseen/Seen/Full in the UC-NF setting.
This means LSG mainly benefits from its LLM supervision instead of refined-query $\bm{Q}^\text{q}_\text{f}$, which requires the HOI detector to provide high-quality attention bias $\bm{A}_\text{f}$.

\end{document}